\documentclass[letterpaper]{article} 
\usepackage{aaai2027}  
\nocopyright
\usepackage[hyphens]{url}  
\usepackage{graphicx} 
\usepackage{natbib}  
\usepackage{caption} 
\usepackage{algorithm}
\usepackage{algorithmic}
\usepackage{amsmath}
\usepackage{amssymb}

\usepackage{newfloat}
\usepackage{listings}
\DeclareCaptionStyle{ruled}{labelfont=normalfont,labelsep=colon,strut=off} 
\floatstyle{ruled}
\newfloat{listing}{tb}{lst}{}
\floatname{listing}{Listing}

\usepackage{booktabs}

\title{Constraint-Bound Agnostic Bayesian Optimization: One Model for All Thresholds}
\author {
    Jin Wang\textsuperscript{\rm 1},
    Xi Lin\textsuperscript{\rm 2},
    Handing Wang\textsuperscript{\rm 1}\corresponding
}
\affiliations {
    \textsuperscript{\rm 1}School of Artificial Intelligence, Xidian University, China\\
    \textsuperscript{\rm 2}School of Mathematics and Statistics, Xi'an Jiaotong University, China\\
    jinwang\_2002@163.com, xi.lin@xjtu.edu.cn, hdwang@xidian.edu.cn
}

\begin{document}

\maketitle

\begin{abstract}
Expensive constrained optimization problems in real-world industry design often involve constraint thresholds that are difficult to determine in advance. Engineers may need to adjust constraint thresholds to explore different feasibility–performance trade-offs, requiring solutions under a wide range of threshold settings. However, existing constrained Bayesian optimization methods treat each threshold configuration independently, leading to repeated optimization and failing to exploit the shared relationship among continuously varying thresholds. To address this challenge, we propose constraint-bound agnostic Bayesian optimization (CBA-BO), a learning-based framework that learns a parametric constraint model mapping thresholds to optimal solutions. Once learned, CBA-BO directly predicts solutions for arbitrary unseen threshold configurations without additional optimization, with a one-step Bayesian optimization refinement further improving solution quality. Experiments on benchmark and engineering problems demonstrate that CBA-BO learns a transferable threshold-solution mapping, enabling efficient prediction and optimization for arbitrary threshold queries. An intent-guided constraint-bound recommendation mechanism is further developed to improve objective performance while satisfying user-specified constraint preferences.
	
\end{abstract}


\section{Introduction}

Many real-world engineering designs rely on computationally intensive simulations for optimization under practical constraints, leading to expensive constrained optimization problems (ECOPs) in which both the objective and constraint functions are black-box and expensive to evaluate \cite{DSI,DSAEA}.

Bayesian optimization (BO) has emerged as an effective framework for solving such problems due to its strong sample efficiency in global optimization of expensive black-box functions \cite{BO1,BO0}. BO typically constructs surrogate models, such as Gaussian processes (GP), to approximate the expensive objective and constraint functions and iteratively selects promising evaluation points through an acquisition function that balances exploration and exploitation. To handle expensive constraints, a variety of constrained BO methods have been developed \cite{cBO_re}, including Expected Improvement (EI) with constraints \cite{cEI}, constrained BO using noisy EI \cite{NEI}, and the trust-region-based SCBO method \cite{SCBO}. These approaches model the objective and constraints using separate GPs and design acquisition functions that balance objective improvement and feasibility, achieving strong performance on ECOPs.

\begin{figure}[!t]
	\centering
	\includegraphics[width=0.48\textwidth]{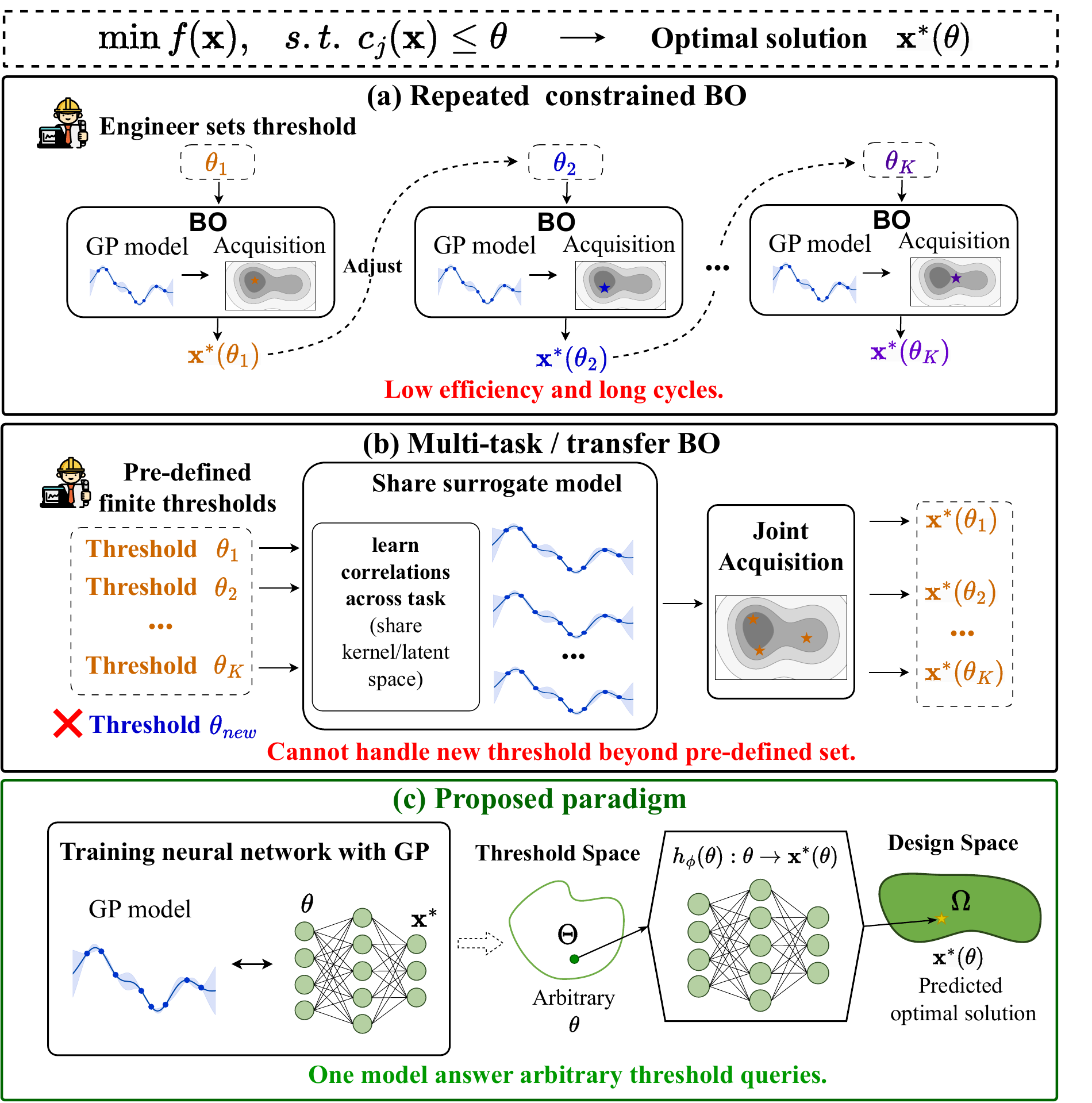} 
	\caption{Architecture comparison of three paradigms for ECOPs with varying constraint-bound thresholds}
	\label{fig1}
\end{figure}

However, constraint-bounds in practice are hard to be determined a priori. Appropriate feasibility thresholds are usually unknown in advance, and engineers often need to repeatedly adjust and test different threshold settings during the design process \cite{gelbart2014bayesian,cBO_re}. Each new threshold configuration effectively defines a new ECOP. Current constrained BO methods treat each threshold configuration independently, requiring optimization to restart whenever the constraint thresholds change.

Since different threshold settings correspond to highly related optimization problems sharing the same underlying objective and constraint functions, it is natural to consider whether knowledge can be transferred across different threshold configurations. Several studies in multi-task and transfer BO \cite{MTBO1,MTBO2} aim to improve the optimization efficiency by sharing information across related tasks. However, these methods are mainly designed for a small finite set of predefined tasks, where the tasks need to be specified before optimization. They transfer information among given tasks with different optimization landscapes, and are not directly applicable to scenarios where constraint thresholds continuously vary or new requirements emerge dynamically.

Instead of transferring information across tasks, another line of research, parametric optimization, studies how optimal solutions vary with problem parameters. Recently, learning-based parametric optimization trains neural networks to map problem parameters to solutions. Examples include unsupervised learning for parametric optimization \cite{nikbakht2020unsupervised}, which uses the Lagrangian as a training loss, and OptINNs \cite{OptINNs}, which embeds KKT conditions into network training. While these methods achieve strong results on parametric optimization problems, they rely on the ability to freely evaluate objective and constraint functions during training. However, this assumption does not hold in the scenario, where the objective and constraint functions are computationally expensive black-box functions. Recently, a few learning-based methods for expensive optimization problems have been proposed. In multi-objective optimization, Pareto set learning (PSL) methods \cite{PSL, cheng2025parametric, cheng2026parametric} combine neural networks with GP models to learn the structure of Pareto-optimal solution manifolds within a unified model, thereby reducing the number of expensive evaluations required to obtain diverse optimal solutions. These studies indicate that learning a shared representation of optimal solutions may provide a promising direction for handling varying constraint-bound requirements in ECOPs.

Therefore, we propose to learn a parametric mapping from constraint thresholds to optimal decisions concurrently with the BO process. Our main contributions include:
\begin{itemize}
	\item A parametric constraint model (PCM) that learns the mapping from constraint thresholds to optimal solutions, enabling direct prediction for arbitrary unseen thresholds without restarting the optimization process.
	
	\item A constraint-bound agnostic Bayesian optimization (CBA-BO) framework that integrates PCM-based solution prediction with GP-guided Bayesian optimization refinement to efficiently exploit learned threshold-solution relationships.
	
	\item Extensive experiments on benchmark and engineering design problems demonstrate the effectiveness and generalization ability of CBA-BO across diverse threshold configurations, with an additional application to intent-guided constraint-bound recommendation.
\end{itemize}

\section{Preliminaries}
\subsection{Expensive Constraint Optimization Problems}

A constrained optimization problem with inequality constraints can be formulated as
\begin{equation}
	\begin{array}{ll}
		\min & f(\mathbf{x}), \ \mathbf{x}=[x_1,x_2,\ldots,x_D] \\
		\mbox{s.t.} & c_j(\mathbf{x})\leq \theta_j,\ j=1,2,\ldots,m
	\end{array}
	\label{eq1}
\end{equation}
where $x_i\in[l_i,u_i]$, $D$ is the number of decision variables, and $m$ is the number of inequality constraints. In ECOPs, both objective and constraint functions are expensive black-box functions, where only function values can be obtained through evaluations.

For a fixed ECOP, constraint thresholds $\theta_j$ are usually incorporated into the standard form $g_j(\mathbf{x})\leq0$, where $g_j(\mathbf{x})=c_j(\mathbf{x})-\theta_j$. However, in practical engineering design, these thresholds are often difficult to determine and may require repeated adjustment. \subsection{Bayesian Optimization} BO is a sample-efficient framework for optimizing expensive black-box functions under limited evaluation budgets. BO iteratively constructs surrogate models from previously evaluated samples and selects new evaluation points through acquisition strategies that balance exploration and exploitation\cite{BO0,BO1}.

\subsubsection{Gaussian Process Model}

GPs are widely used in BO as probabilistic surrogate models for expensive black-box functions. A GP defines a prior distribution over the function space as
\begin{equation}
	f(\mathbf{x})\sim\mathcal{GP}\left(\mu(\mathbf{x}),k(\mathbf{x},\mathbf{x}')\right),
\end{equation}
where $\mu(\mathbf{x})$ and $k(\mathbf{x},\mathbf{x}')$ denote the mean and covariance functions, respectively.

Given $n$ evaluated samples $\mathcal D_n=\{(\mathbf{x}^{(i)},y^{(i)})\}_{i=1}^{n}$, the posterior predictive distribution for a new solution $\mathbf{x}$ follows $f(\mathbf{x})\mid\mathcal D_n\sim\mathcal N\left(\hat\mu(\mathbf{x}),\hat\sigma^2(\mathbf{x})\right),$ where the predictive mean and variance are
\begin{equation}
	\hat\mu(\mathbf{x})=\mu(\mathbf{x})+\mathbf{k}^\top\mathbf{K}^{-1}\mathbf y,
\end{equation}
\begin{equation}
	\hat\sigma^2(\mathbf{x})=k(\mathbf{x},\mathbf{x})-\mathbf{k}^\top\mathbf{K}^{-1}\mathbf{k},
\end{equation}
with $\mathbf{k}=k(\mathbf{x},X)$ and $\mathbf{K}=k(X,X)$. The GP hyperparameters are estimated by maximizing the marginal likelihood.

\subsubsection{Log Constrained Expected Improvement}
Expected improvement (EI) is a widely used acquisition strategy in BO. For constrained optimization, constrained expected improvement (cEI) incorporates feasibility by weighting EI with the probability of satisfying the constraints \cite{cEI}. However, cEI may suffer from numerical instability when feasibility probabilities become extremely small. To address this issue, log-constrained expected improvement (logcEI) reformulates the acquisition function in logarithmic space, improving numerical stability and optimization performance in constrained BO \cite{logEI}. logcEI can be expressed as
\begin{equation} LogcEI ( \mathbf{x} ) = \log \left( EI ( \mathbf{x} ) \right) + \sum_{j=1}^{m} \log P \left( c_j ( \mathbf{x} ) \le \theta_j \right), \end{equation}
where $\mathrm{EI}(\mathbf{x})$ denotes the expected improvement and $P(c_j(\mathbf{x})\le \theta_j)$ is the feasibility probability of the $j$-th constraint.

\section{Proposed Method}
For a threshold vector $\boldsymbol{\theta} = [\theta_1,\ldots,\theta_m]^\top$, Equation~(\ref{eq1}) defines an ECOP. Varying $\boldsymbol{\theta}$ within a threshold domain $\Theta = [\boldsymbol{\theta}^{L}, \boldsymbol{\theta}^{U}]$ defines a family of related ECOPs sharing the same objective and constraint functions, where $\boldsymbol{\theta}^{L}$ and $\boldsymbol{\theta}^{U}$ denote the lower and upper bounds, respectively. For each $\boldsymbol{\theta}\in\Theta$, the corresponding optimal solution is
\begin{equation}
	\mathbf{x}^{*}(\boldsymbol{\theta})
	= \arg\min_{\mathbf{x}} f(\mathbf{x})
	\quad \text{s.t.} \quad
	c_j(\mathbf{x}) \le \theta_j, \quad j=1,\ldots,m.
\end{equation}

Conventional constrained BO typically treats each constraint-threshold configuration as an independent optimization problem. When feasibility requirements vary continuously, repeatedly solving individual ECOP instances becomes increasingly inefficient and limits knowledge sharing across related problems. To address this issue, we propose \emph{Constraint-Bound Agnostic Bayesian Optimization} (CBA-BO), a framework that learns a shared parametric constraint model (PCM) across varying threshold settings. Instead of independently optimizing each ECOP instance, CBA-BO directly models the relationship between threshold vectors and their corresponding optimal solutions, thereby enabling transferable optimization over a family of related ECOPs.

\begin{algorithm}[ht]
	\caption{Constraint-Bound Agnostic Bayesian Optimization (CBA-BO)}
	\label{alg:CBABO}
	\begin{algorithmic}[1]
		\STATE Initialize database $\mathcal{D}_0$ using LHS
		\STATE Initialize parametric constraint model $h_{\phi}$
		\FOR{$t = 1$ to $T$}
		\STATE Fit GP surrogate models using $\mathcal{D}_{t-1}$
		\FOR{$s = 1$ to $S$}
		\STATE Sample threshold vectors $\boldsymbol{\theta}\sim\Theta$
		\STATE Update $h_{\phi}$ using GP-guided gradient learning
		\ENDFOR
		\STATE Generate candidate set
		$\mathcal{X}_t = \{h_{\phi}(\boldsymbol{\theta})\mid\boldsymbol{\theta}\sim\Theta\}$
		\STATE Select batch $\mathcal{B}_t$ using sampling strategy
		\STATE Evaluate $\mathcal{B}_t$ and update $\mathcal{D}_t\leftarrow \mathcal{D}_{t-1}\cup\mathcal{B}_t$
		\ENDFOR
		\RETURN
		$h_{\phi}$ and $\mathcal{D}_T$
	\end{algorithmic}
\end{algorithm}

Algorithm~\ref{alg:CBABO} summarizes the workflow of CBA-BO. The framework integrates a parametric constraint model (PCM), GP-based surrogate guidance, and online data acquisition into a unified optimization process. Specifically, the PCM $h_{\phi}$ is trained to learn the relationship between constraint thresholds and optimal solutions. GPs are constructed from evaluated samples to estimate objective values and constraint satisfaction, providing uncertainty-aware guidance for updating the PCM through gradient-based optimization. To improve sample efficiency, threshold candidates generated by the PCM are
evaluated by the surrogate models, and a diverse batch is selected for expensive evaluations. Through iterative updates of both surrogate models and the parametric generator, CBA-BO learns a reusable solution model that can generalize to arbitrary unseen threshold settings.

\subsection{Parametric Constraint Model}

To model the dependence of optimal solutions on threshold settings, we construct a PCM 
\begin{equation}
	\mathbf{x}=h_{\phi}(\boldsymbol{\theta}),
\end{equation}
where $\boldsymbol{\theta}\in\Theta$ is the threshold vector, $\mathbf{x}\in\Omega$ is the generated optimal solution, and $h_{\phi}$ is a neural network model parameterized by $\phi$.

The PCM aims to approximate the relationship between threshold vectors and their corresponding optimal solutions. For numerical stability and consistent model training, both $\boldsymbol{\theta}$ and $\mathbf{x}$ are normalized into the unit hypercube $[0,1]^m$ and $[0,1]^D$, respectively. In this work, the PCM $h_{\phi}(\boldsymbol{\theta})$ is implemented as an MLP neural network, which is good at capturing complicated problem structures \cite{sener2020learning}. The input and output dimensions correspond to the number of constraints $m$ and decision variables $D$, respectively. The model details can be found in Appendix.

\subsection{GP-Guided Parametric Constraint Learning}

Given the PCM $h_{\phi}(\boldsymbol{\theta})$, our goal is to learn the model parameter$\phi$ such that the generated solution $\mathbf{x}$ approximates the optimal solution corresponding to arbitrary threshold settings within $\Theta$. Denote the generated solution set as $\mathcal{X}_h=\{h_{\phi}(\boldsymbol{\theta})\mid\boldsymbol{\theta}\in\Theta\}.$

Ideally, the learned model should satisfy
\begin{equation}
	h_{\phi^*}(\boldsymbol{\theta})=\mathbf{x}^*(\boldsymbol{\theta}),\quad\forall\boldsymbol{\theta}\in\Theta,
\end{equation}
where
$\mathbf{x}^*(\boldsymbol{\theta})$ denotes the optimal solution under threshold vector $\boldsymbol{\theta}$. However, since the true optimal mapping is unknown and expensive to obtain, we instead learn the model using GP surrogate guidance. For a threshold vector $\boldsymbol{\theta}$, the generated solution $\mathbf{x}=h_{\phi}(\boldsymbol{\theta})$ should simultaneously satisfy feasibility requirements and achieve good objective quality. To this end, we construct a feasibility-aware surrogate objective.

For minimization problems, the objective surrogate is defined using the lower confidence bound (LCB) \cite{LCB},
\begin{equation} \hat f ( \mathbf{x} ) = \mathrm{LCB}_f ( \mathbf{x} ) = \hat\mu_f ( \mathbf{x} ) - \beta \hat\sigma_f ( \mathbf{x} ), \end{equation}
where $\hat\mu_f$ and $\hat\sigma_f$ are the GP posterior mean and standard deviation, respectively.

Constraint violation under threshold $\boldsymbol{\theta}$ is estimated conservatively using upper confidence bounds (UCB) \cite{LCB},
\begin{equation} \hat v_j ( \mathbf{x} ) = \max \left( 0, \mathrm{UCB}_{c_j} ( \mathbf{x} ) - \theta_j \right), \end{equation}
where $\mathrm{UCB}_{c_j}(\mathbf{x})=\hat\mu_{c_j}(\mathbf{x})+\beta\hat\sigma_{c_j}(\mathbf{x}).$

A generated solution is regarded as feasible if $v_j(\mathbf{x})=0$ for all constraints. Accordingly, we define the feasibility-aware surrogate objective as
\begin{equation} \hat z ( \mathbf{x} | \boldsymbol{\theta} ) = \begin{cases} \hat f ( \mathbf{x} ), & \text{if feasible}, \\ \sum_{j=1}^{m} \hat v_j ( \mathbf{x} ), & \text{otherwise}. \end{cases} \end{equation}

The PCM is learned by minimizing the expected surrogate objective over the threshold domain,
\begin{equation} \phi^* = \arg \min_{\phi} \; \mathbf{E}_{\boldsymbol{\theta} \sim \Theta} \left[ \hat z \left( h_{\phi} ( \boldsymbol{\theta} ) | \boldsymbol{\theta} \right) \right]. \end{equation}

The expectation over continuously varying threshold vectors is generally intractable. Therefore, we employ threshold batch sampling and gradient descent for model learning,
\begin{equation} \phi_{s+1} = \phi_s - \eta \frac{1}{K} \sum_{k=1}^{K} \nabla_{\phi} \hat z \left( h_{\phi} ( \boldsymbol{\theta}_k ) | \boldsymbol{\theta}_k \right), \end{equation}
where $ \{\boldsymbol{\theta}_k\}_{k=1}^{K}\sim\Theta$ are generated by a structured mixture sampling strategy over $\Theta$, consisting of diagonal uniform interpolation, diagonal stochastic sampling, and Sobol quasi-random global sampling, and $\eta$ is the learning rate.

The model update follows the chain rule
$
	\nabla_{\phi}\hat z=\frac{\partial\hat z}{\partial
	\mathbf{x}}\frac{\partial\mathbf{x}}{\partial\phi},
$
where $\partial\hat z/\partial\mathbf{x}$ is obtained from GP surrogate gradients and $\partial\mathbf{x}/\partial\phi$ is computed through automatic differentiation of the neural network $h_{\phi}$. For feasible solutions, $\partial\hat z/\partial\mathbf{x}$ corresponds to the gradient of $\mathrm{LCB}_f$, whereas infeasible solutions are guided toward feasibility using gradients of violated constraints.

\paragraph{Monotonic Regularization.}
Relaxing feasibility thresholds generally enlarges the feasible region of a minimization problem. Therefore, the optimal objective value should not increase when constraint thresholds become relaxed, since previously feasible solutions remain feasible. This monotonic relationship provides
useful prior knowledge for learning the threshold solution mapping. However, due to limited expensive evaluations and surrogate approximation errors, the learned parametric model may produce inconsistent responses for different threshold settings. To enforce such threshold response consistency, we introduce monotonic guidance into the surrogate objective in the second half of the training process.

Using the chain rule positive sensitivity of objective quality to threshold relaxation is penalized through
\begin{equation}
	r_{mono}=
	\mathbf{E}\left[\max\left(0,\frac{\partial \hat f}
	{\partial\mathbf{x}}\frac{\partial\mathbf{x}}{\partial\boldsymbol{\theta}}
	\right)\right].
\end{equation}

Accordingly, the surrogate objective is augmented as
\begin{equation}
	\hat z_{aug}(\mathbf{x}|\boldsymbol{\theta})=\hat z
	(\mathbf{x}|\boldsymbol{\theta})+\alpha r_{mono},
\end{equation}
where
$
\alpha
$
controls the influence of monotonic guidance. The PCM is therefore learned by minimizing the expected augmented surrogate objective.

\subsection{Sampling Strategy}

After training the parametric model, a set of threshold vectors is sampled from the threshold space and mapped to corresponding candidate solutions,
\begin{equation}
\mathcal{X}_t = \{ h_{\phi}(\boldsymbol{\theta}) \mid
\boldsymbol{\theta} \sim \Theta \}.
\end{equation}
The generated candidates are evaluated using the Gaussian process surrogate, and a batch $\mathcal{B}_t$ of five candidates is selected considering predicted performance and solution diversity. The selected batch includes two boundary cases with the lower and upper threshold limits and three randomly sampled intermediate configurations. The selected solutions are then evaluated using the expensive objective and constraint functions. For each evaluated sample, the threshold vector and its corresponding solution are added to the training database $\mathcal{D}_t$ to update the parametric constraint model. The model details can be found in Appendix.

\subsection{Local Refinement}

Although $h_{\phi}$ generates high-quality solutions rapidly, prediction errors may still arise due to limited surrogate accuracy and imperfect model generalization across the threshold domain. To further improve solution reliability, we introduce a one-step local refinement stage.

Given the predicted solution $\hat{\mathbf{x}}^{*} = h_{\phi}(\boldsymbol{\theta})$ for any queried threshold vector $\boldsymbol{\theta}$, we construct a local design region centered at $\hat{\mathbf{x}}^{*}$,
$
	\Omega_{local} = [\hat{\mathbf{x}}^{*} - r,\; \hat{\mathbf{x}}^{*} + r],
$ where $r$ is a fixed fraction of the global search domain width. Within $\Omega_{local}$, the collected database $\mathcal{D}_T$ is reused to refit GP surrogate models, avoiding additional expensive evaluations for initialization.

To balance objective improvement and feasibility, LogcEI is employed as the acquisition function. The local refinement performs $T_{local} = 1$ sequential evaluations,
\begin{equation}
	\mathbf{x}_{t+1} =
	\arg\max_{\mathbf{x} \in \Omega_{local}}\;
	\alpha_{\mathrm{LogcEI}}(\mathbf{x}),
\end{equation}
and the candidate with the highest $\alpha_{\mathrm{LogcEI}}$ value among all evaluated points is returned as the refined solution for $\boldsymbol{\theta}$.

\subsection{Constraint-Bound Recommendation}

Once $h_{\phi}$ is learned, it provides a reusable mapping from constraint thresholds to corresponding solutions, allowing engineers to efficiently explore a large number of possible constraint configurations without repeatedly solving new ECOP instances. In practical engineering design, constraint thresholds are often adjusted iteratively through trial and error to balance feasibility and performance, which can be time-consuming and highly dependent on expert experience. By leveraging the learned PCM, constraint-bound recommendation aims to accelerate this exploration process by identifying promising threshold configurations according to engineers' preferences. Given a baseline threshold $\boldsymbol{\theta}^{base}$ and user-specified adjustment intentions $\{a_j\}_{j=1}^{m}$ where $a_j \in \{-1, 0, 1\}$ denotes tighten, lock, loosen respectively, the mechanism identifies a new threshold vector $\boldsymbol{\theta}^{rec}$ that improves objective performance while respecting the engineer's intentions on each constraint bound.

A local threshold set $\Theta_{local}$ is generated around $\boldsymbol{\theta}^{base}$ in the normalized threshold space $[0,1]^m$. To balance intent following and local exploration, the candidate set is composed of an intent-guided subset and an intent-independent exploration subset $\Theta_{local} = \Theta_{intent} \cup \Theta_{explore}. $ For the intent-guided subset, each dimension $j$ is perturbed independently as $\theta_j^{(i)}=\theta_j^{base}+\delta_j^{(i)}$, where the perturbation $\delta_j^{(i)}$ is sampled according to the intent level: \begin{equation} \delta_j^{(i)} \sim \begin{cases} -|\mathcal{N}(0,\, 0.5r_j)| & a_j = -1 \quad \text{(tighten)}, \\ \mathcal{U}(-0.025r_j,\, 0.025r_j) & a_j = 0 \quad \text{(lock)}, \\ |\mathcal{N}(0,\, 0.5r_j)| & a_j = 1 \quad \text{(loosen)}. \end{cases} \end{equation} For the intent-independent exploration subset, unbiased Gaussian perturbations are sampled around the baseline threshold to preserve local exploration. All sampled threshold vectors are clamped to $[0,1]^m$.

For each $\boldsymbol{\theta}^{(i)} \in \Theta_{local}$, the learned model generates $\hat{\mathbf{x}}^{*(i)} = h_{\phi}(\boldsymbol{\theta}^{(i)})$, which is retained only if GP mean predictions confirm feasibility,
$
\hat{\mu}_{c_j}(\hat{\mathbf{x}}^{*(i)}) \le \theta_j^{(i)},~ \forall j.
$ Feasible candidates are ranked by a utility score balancing objective improvement and adjustment cost,
\begin{equation}
	S(\boldsymbol{\theta}^{(i)}) = w_f \Delta f
	+ R(\boldsymbol{\theta}^{(i)}) - P(\boldsymbol{\theta}^{(i)}),
\end{equation}
where
$
\Delta f = \hat{\mu}_f(\hat{\mathbf{x}}^{*base}) -
\hat{\mu}_f(\hat{\mathbf{x}}^{*(i)})
$ is the predicted objective improvement over the baseline,
\begin{equation}
	R(\boldsymbol{\theta}^{(i)}) = \lambda_r \sum_{j:\, a_j=-1}
	\max(0,\, -\delta_j^{(i)})
\end{equation}
rewards tightening adjustments proportional to their magnitude, and
\begin{equation}
	P(\boldsymbol{\theta}^{(i)}) = \sum_{j=1}^{m}
	\begin{cases}
		\lambda_h \max(0,\, \delta_j^{(i)}) & a_j \in \{-1, 0\}, \\
		\lambda_r \max(0,\, \delta_j^{(i)} - \epsilon_j)^2 & a_j = 1,
	\end{cases}
\end{equation}
penalizes relaxation beyond the baseline, where for $a_j \in \{-1, 0\}$ any positive $\delta_j^{(i)}$ is heavily penalized to enforce a hard barrier against intent-violating relaxation, and for $a_j = 1$ only relaxation exceeding the allowance $\epsilon_j$ is penalized. The recommended configuration is selected as
\begin{equation}
	\boldsymbol{\theta}^{rec} =
	\arg\max_{\boldsymbol{\theta}^{(i)} \in \Theta_{local}}
	S(\boldsymbol{\theta}^{(i)}).
\end{equation}

\section{Experiments}

\subsubsection{Baseline Algorithms.}

To the best of our knowledge, there are currently no constrained optimization methods specifically developed for simultaneously solving multiple ECOPs induced by continuously varying constraint-threshold settings. Therefore, we compare the proposed CBA-BO with several representative constrained optimization methods, including CMA-ES \cite{cmaes,cmaes1}, cEI \cite{cEI}, ALBO \cite{ALBO}, SCBO \cite{SCBO}, and LogcEI \cite{logEI}. These methods cover both evolutionary optimization and state-of-the-art constrained Bayesian optimization approaches. Since all baseline methods are designed for a single threshold configuration, each threshold setting is treated as an independent ECOP and optimized separately in the experiments. The BO-based baselines are implemented using BoTorch~\cite{balandat2020botorch}, with their parameter settings following the corresponding original papers whenever applicable.

\subsubsection{Benchmarks and Real-World Problems}

Experiments are conducted on sixteen constrained optimization problems, including eleven synthetic benchmarks and five engineering design problems. The synthetic benchmarks include problems from CEC 2006 and CEC 2010 \cite{CEC2006,CEC2010}, Ackley functions \cite{SCBO}, and constrained BO benchmarks such as GKXWC1, GKXWC2 \cite{cEI}, and Keane Bump \cite{keane}. The engineering problems include pressure vessel design \cite{Press}, speed reducer design, reinforced concrete beam design, three-bar truss design \cite{RCBeam}, and car side impact design \cite{car,probs}. Detailed problem descriptions and threshold domains are provided in the Appendix.

\begin{figure*}[!ht]
	\centering
	\includegraphics[width=1\textwidth]{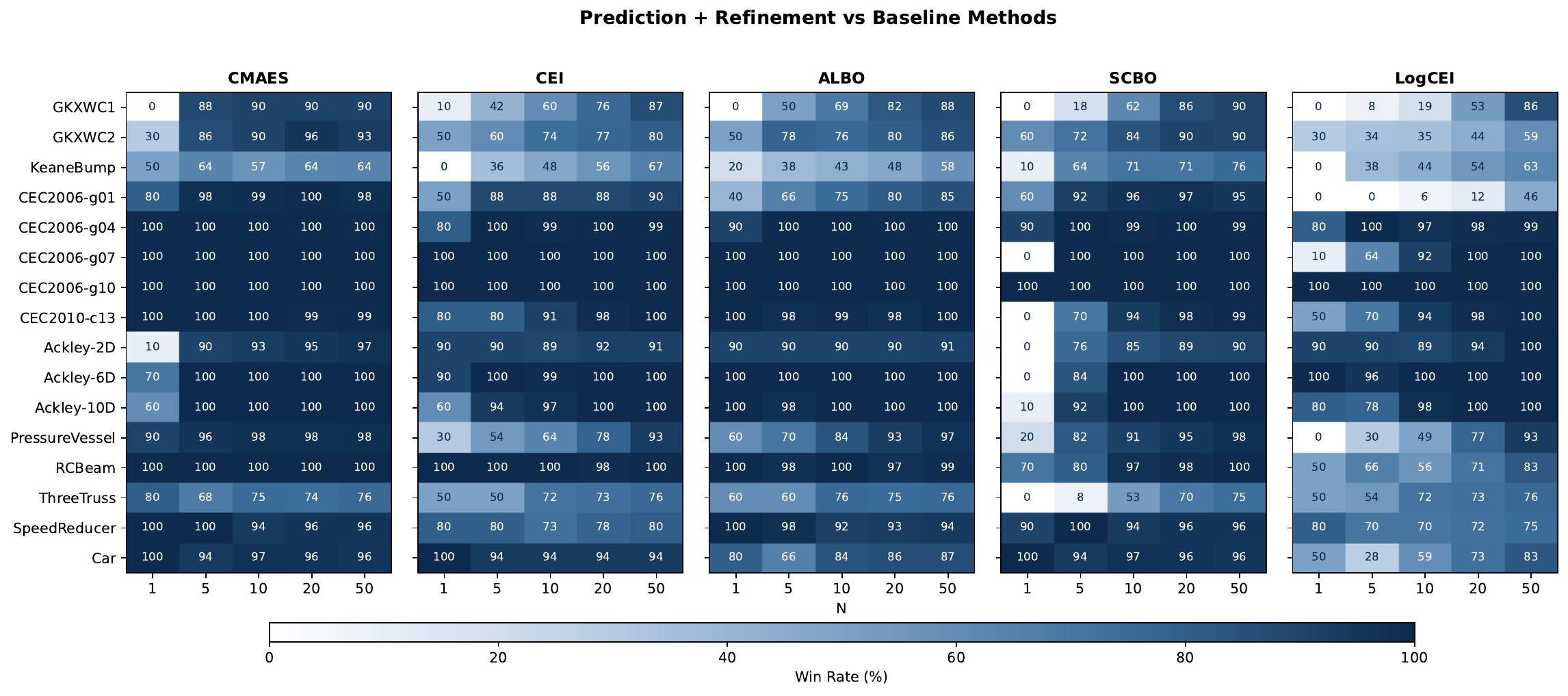} 
	\caption{Win rate (\%) of CBA-BO with one-step refinement against baseline methods under different numbers of threshold configurations. Rows denote benchmark problems and columns denote the number of threshold configurations.}
	\label{fig2}
\end{figure*}
\subsubsection{Experimental Settings}

For all problems, 30 LHS initial sample setting are shared by all algorithms. The implementation details and parameter settings of the CBA-BO are provided in the Appendix.

Since existing constrained BO baselines are designed for a single threshold vector, we introduce different numbers of queried threshold configurations $N$ only as a unified evaluation protocol. The value of $N$ does not limit CBA-BO, which can generate solutions for arbitrary threshold queries after training. For each $N$, the same threshold group is used by all algorithms. Since the baselines are designed for a single threshold configuration, each threshold vector is optimized independently. After sharing the 30 initial samples, the remaining 170 evaluations are evenly distributed among the $N$ ECOP instances, and two additional evaluations are assigned to each threshold configuration for fair comparison with CBA-BO refinement.

For CBA-BO, we report two settings. In \emph{Direct Prediction}, the learned model predicts one solution for each threshold vector, which is then evaluated once by the true black-box functions. In \emph{Refinement}, one additional local BO evaluation is performed after evaluating the predicted solution. All algorithms are run independently 10 times. For each run and each $N$, a new threshold group is randomly generated and shared by all algorithms. Win rate is computed under the same threshold vector: feasible solutions are preferred; if both are feasible, objective values are compared; otherwise, total constraint violations are compared. Metrics are averaged over all thresholds and 10 runs. The runtime analysis is provided in the Appendix.

To further examine scalability, we additionally consider a setting with $N=1000$ queried threshold configurations. Since CBA-BO generates solutions through a shared parametric model, increasing the number of threshold configurations does not require retraining or repeated optimization. Conventional optimization algorithms are excluded from this setting because solving 1,000 ECOPs independently would be computationally prohibitive.

\subsection{Experimental Results and Analysis}

\begin{table}[h]
	\centering
	\caption{Feasibility ratio (\%) of direct prediction and one-step refinement over different test problems ($N=1000$).}
	\label{tab:feasibility}
	\renewcommand{\arraystretch}{1}
    \small
	\setlength{\tabcolsep}{12pt}

	\begin{tabular}{lcc}
		\hline
		Problem & Prediction & Refinement \\
		\hline
		CEC2006-g01   & $39.5 \pm 1.1$   & $\mathbf{99.5 \pm 0.2}$ \\
		CEC2006-g04   & $97.6 \pm 0.4$   & $\mathbf{98.7 \pm 1.1}$ \\
		CEC2006-g07   & $69.2 \pm 5.5$   & $\mathbf{100.0 \pm 0.1}$ \\
		CEC2006-g10   & $99.1 \pm 0.8$   & $\mathbf{100.0 \pm 0.1}$ \\
		CEC2010-c13   & $79.1 \pm 13.8$  & $\mathbf{99.7 \pm 0.2}$ \\
		Ackley-2D     & $100.0 \pm 0.2$  & $\mathbf{100.0 \pm 0.0}$ \\
		Ackley-6D     & $99.9 \pm 0.2$   & $\mathbf{100.0 \pm 0.0}$ \\
		Ackley-10D    & $99.8 \pm 0.2$   & $\mathbf{100.0 \pm 0.0}$ \\
		GKXWC1        & $100.0 \pm 0.0$  & $\mathbf{100.0 \pm 0.0}$ \\
		GKXWC2        & $90.5 \pm 0.5$   & $\mathbf{99.8 \pm 0.3}$ \\
		KeaneBump     & $92.5 \pm 15.8$  & $92.5 \pm 15.8$ \\
		PressureVessel& $87.7 \pm 3.0$   & $\mathbf{100.0 \pm 0.0}$ \\
		RCBeam        & $81.9 \pm 1.3$   & $\mathbf{100.0 \pm 0.1}$ \\
		ThreeTruss    & $68.2 \pm 6.7$   & $\mathbf{91.4 \pm 4.0}$ \\
		SpeedReducer  & $91.7 \pm 0.1$   & $\mathbf{96.8 \pm 0.1}$ \\
		Car           & $72.1 \pm 0.9$   & $\mathbf{98.3 \pm 0.3}$ \\
		\hline
	\end{tabular}
\end{table}

Figure ~\ref{fig2} reports the average win rate of CBA-BO with one-step refinement against five baselines under different numbers of threshold configurations. A higher win rate means that CBA-BO obtains better solutions on a larger proportion of threshold settings. Overall, CBA-BO achieves competitive performance on both benchmark and engineering problems. It consistently outperforms CMA-ES, cEI and ALBO on most problems, showing that the learned threshold-solution mapping uses expensive evaluations more effectively than independent evolutionary search or augmented-Lagrangian BO. Compared with SCBO and LogcEI, CBA-BO also achieves favorable results on many problems, especially when multiple threshold configurations are considered.

The most important trend in Figure ~\ref{fig2} is that the advantage of CBA-BO becomes more pronounced as the number of threshold configurations increases. When $N=1$, dedicated BO methods such as SCBO and LogcEI can be highly competitive because the entire evaluation budget is devoted to a single ECOP. In particular, LogcEI performs sequential, threshold-specific acquisition optimization, which can be very effective on problems where a single constrained optimum can be located with sufficient evaluations. Therefore, CBA-BO does not always dominate LogcEI on some problems with relatively regular feasible regions or strong local acquisition guidance. However, as $N$ increases, the per-threshold budget of all baseline methods decreases because they solve each threshold configuration independently. In contrast, CBA-BO learns a shared parametric model over the threshold space and reuses information across related ECOPs. As a result, the cost of model learning is amortized over multiple threshold configurations, leading to increasingly higher win rates when $N$ grows from $1$ to $50$.

This trend is clear on CEC2006-g01, CEC2010-c13, Pressure Vessel, Three-Truss, and Car, where the win rates against strong BO baselines increase substantially with more threshold configurations. These results support the motivation of CBA-BO: although dedicated constrained BO methods can be effective for individual ECOPs, they require repeatedly solving new optimization problems when constraint thresholds change. By learning a unified parametric mapping from constraint thresholds to solutions, CBA-BO can efficiently generalize to continuously arbitrary unseen threshold queries. The detailed objective values and the effects of PCM prediction and refinement are provided in the Appendix.

To further demonstrate the query efficiency of the learned parametric mapping, we additionally evaluate CBA-BO on 1000 randomly sampled threshold queries. Table~\ref{tab:feasibility} further evaluates scalability under a setting with $N=1000$ threshold configurations. Direct prediction already achieves high feasibility on several problems, indicating that the learned model captures useful global structure in the threshold-solution relationship. Nevertheless, prediction errors remain on problems with complex or narrow feasible regions. With only one additional expensive evaluation, the refinement step substantially improves feasibility. For example, the feasibility ratio increases from $39.53\%$ to $99.54\%$ on CEC2006-g01, from $69.16\%$ to $99.95\%$ on CEC2006-g07, and from $72.12\%$ to $98.31\%$ on the Car problem. This confirms that CBA-BO provides reliable initial solutions, while local BO refinement corrects remaining local prediction errors. Additional visualizations in the Appendix show how constraint activity affects PCM prediction and how local refinement corrects errors. The contribution of PCM-based initialization to the refinement process is further analyzed in the Appendix.

\subsection{Ablation Study}
\label{sec:ablation}

We validate the contribution of each component by comparing the full
CBA-BO with two variants: (1) \textbf{All Rand} -- replacing the PCM-generated candidate solutions with randomly sampled candidates, thereby
removing the guidance provided by the learned threshold--solution mapping and (2) \textbf{No Mono} -- removing the monotonic guidance during PCM
training. The ablation study is conducted on five representative problems. For each problem and each variant, we perform five independent runs. In each run, 1000 test threshold configurations are sampled, and the direct prediction results of full CBA-BO and the corresponding variant are compared to compute the win rate.

\begin{table}[h]
	\centering
	\caption{Ablation study: win rate (\%) of full CBA-BO against each variant.}
	\label{tab:ablation}
	\renewcommand{\arraystretch}{1.2}
	\small
	\begin{tabular}{lcccccc}
		\hline
		Variant & g7 & g10 & c13 & PressV & ThreeT& Avg \\
		\hline
		All Rand  & 50.0 & 99.3 & 97.1 & 94.3 & 76.8 & 83.5 \\
		No Mono   & 63.5 & 93.9 & 68.8 & 54.2 & 61.6 & 68.4 \\
		\hline
	\end{tabular}
\end{table}

The results in Table~\ref{tab:ablation} demonstrate that PCM plays a critical role in CBA-BO by providing informative prior solutions for guiding the Bayesian optimization process. Removing PCM leads to clear performance degradation, indicating that the learned threshold--solution relationship helps generate high-quality initial candidates, improve surrogate modeling efficiency, and guide the search toward promising regions under varying constraint configurations.The removal of monotonic guidance results in relatively smaller changes, suggesting that its effect is not mainly reflected in average optimization performance. Instead, monotonic guidance provides additional structural regularization by encouraging more consistent responses of the parametric model under different threshold settings. A extra analysis of this effect is provided in Appendix.

\subsection{Case Study: Intent-Guided Constraint-Bound Recommendation}
\label{sec:case}

\begin{table}[h]
	\centering
	\caption{Recommended constraint thresholds under mixed intents.}
    \small
	\begin{tabular}{ccccr}
		\hline
		Constraint & Intent & $\theta^{base}$ & $\theta^{rec}$ & Real $c$ \\
		\hline
		$c_1$ & tighten & 3.0 & 0.51 & $-0.31$ \\
		$c_2$ & lock    & 3.0 & 2.99 & $-0.65$ \\
		$c_3$ & tighten & 3.0 & 2.63 & $-0.67$ \\
		$c_4$ & lock    & 3.0 & 3.00 & $-0.19$ \\
		$c_5$ & loosen  & 3.0 & 3.44 & $-2.70$ \\
		$c_6$ & lock    & 3.0 & 2.98 & $-2.82$ \\
		$c_7$ & loosen  & 3.0 & 5.84 & $5.81$ \\
		$c_8$ & loosen  & 3.0 & 3.52 & $0.26$ \\
		$c_9$ & tighten & 3.0 & 1.94 & $-0.70$ \\
		$c_{10}$ & lock & 3.0 & 2.97 & $-0.26$ \\
		\hline
		\multicolumn{5}{c}{Objective $f$: $17.60 \rightarrow 15.70$ ($-10.7\%$)} \\
		\multicolumn{5}{c}{Feasibility: $c_i(\mathbf{x}^{rec}) \leq \theta_i^{rec}$ for all constraints} \\
		\hline
	\end{tabular}
	\label{tab}
\end{table}

We demonstrate the intent-guided constraint-bound recommendation mechanism on the Car problem, which contains 11 decision variables and 10 constraints. Starting from a moderate baseline setting $\theta_i^{base}=3.0$ for all constraints, we specify tightening intents for $c_1$, $c_3$, and $c_9$, loosening intents for $c_5$, $c_7$, and $c_8$, and lock the remaining four constraints.

As shown in Table~\ref{tab}, the recommended thresholds are consistent with the specified intents. The tightened thresholds are reduced by $12$--$83\%$, the loosened thresholds are increased by $15$--$95\%$, and the locked thresholds remain almost unchanged. Although some raw constraint values are positive, the returned design is feasible under the recommended thresholds, i.e., $c_i(\mathbf{x}) \leq \theta_i^{rec}$ for all constraints. By relaxing non-critical constraints, the system identifies a design with $f=15.70$, corresponding to a $10.7\%$ improvement over the baseline objective value. This case study shows that the proposed recommendation mechanism enables users to explore the constraint--objective trade-off through intuitive high-level intents, without repeated manual tuning of a high-dimensional threshold vector.

\section{Conclusions}

This paper proposed CBA-BO, a constraint-bound-agnostic Bayesian optimization framework for expensive constrained optimization with varying feasibility thresholds. By learning a shared parametric mapping from constraint thresholds to optimal solutions, CBA-BO can directly provide high-quality solutions for arbitrary unseen threshold configurations without restarting optimization. The PCM is trained through a GP-guided optimization process to capture the threshold-solution relationship, while one-step local refinement further improves solution quality when direct predictions are insufficient.  Experiments on benchmark and engineering design problems demonstrate the effectiveness of CBA-BO in achieving high-quality solutions while handling varying constraint bounds. The learned parametric model also provides a practical basis for constraint-bound recommendation and feasibility--performance trade-off exploration.

Several limitations remain. The quality of the learned threshold--solution mapping depends on the accuracy of GP surrogate modeling, which may degrade in high-dimensional problems or highly irregular feasible regions. Moreover, the current recommendation mechanism only supports simple high-level adjustment intents. Future work will investigate more expressive preference modeling and extend CBA-BO to more complex real-world constrained design scenarios.

\bibliography{aaai2027}

\newpage
\section{Benchmark Problem Formulations}

This section summarizes the benchmark problems used in the experiments. For all problems, the objective and constraint functions follow their standard formulations in the corresponding original references. The threshold domains used in our experiments are reported in Table~\ref{tab:problem_summary}. Since CBA-BO studies varying constraint thresholds, we keep the original objective and constraint functions unchanged and only vary the feasibility thresholds $\boldsymbol{\theta}$ within the specified domains. \subsection{Engineering Design Problems} The engineering design problems used in the experiments are briefly described as follows.
\begin{itemize} 
	\item \textbf{Three-Truss Design.} This problem aims to minimize the structural volume of a three-bar truss. The design variables describe the cross-sectional areas of the bars, and the constraints are imposed by stress limits on different truss members. The formulation follows the standard three-bar truss design problem used in structural optimization \cite{RCBeam}. 
	
	\item \textbf{Reinforced Concrete Beam Design.} This problem minimizes the construction cost of a reinforced concrete beam. The design involves both discrete and continuous variables, including the reinforcing bar area, beam width, and beam depth. The constraints are related to load-bearing capacity and safety requirements. The formulation follows the reinforced concrete beam design benchmark in \cite{RCBeam}.
	
	\item \textbf{Pressure Vessel Design.} This problem seeks to minimize the manufacturing cost of a cylindrical pressure vessel. The decision variables include the shell thickness, head thickness, inner radius, and cylindrical length. The constraints are associated with pressure-vessel design requirements such as thickness, volume, and length limitations \cite{PressV}. The formulation is taken from \cite{Press}.
	
	\item \textbf{Speed Reducer Design.} This problem minimizes the weight of a mechanical speed reducer. The decision variables describe key gear and shaft dimensions, and the constraints involve gear geometry, shaft stresses, and mechanical design restrictions. The formulation follows the commonly used speed reducer design benchmark in engineering optimization \cite{RCBeam}. 
	
	\item \textbf{Car Side Impact Design.} This problem aims to minimize vehicle weight while satisfying crashworthiness requirements. The design variables describe structural components and reinforcement settings of the car body, and the constraints are related to impact safety and compatibility requirements. The formulation is taken from the car side impact design problem in \cite{car,probs}. 
	
\end{itemize}

\begin{table}[h]
	\centering
	\caption{Summary of the test problems and threshold domains used in the experiments. Here $a^{\times k}$ denotes that value $a$ is repeated $k$ times.}
	\label{tab:problem_summary}
	\renewcommand{\arraystretch}{1.2}
	\setlength{\tabcolsep}{5pt}
	\begin{tabular}{lcccc}
		\hline
		Problem & $D$ & $m$ & $\boldsymbol{\theta}^{L}$ & $\boldsymbol{\theta}^{U}$ \\
		\hline
		CEC2006-g01 & 13 & 9  & $[0^{\times9}]$  & $[210^{\times3},110^{\times6}]$ \\
		CEC2006-g04 & 5  & 6  & $[0^{\times6}]$  & $[500^{\times3},110^{\times3}]$ \\
		CEC2006-g07 & 10 & 8  & $[0^{\times8}]$  & $[300^{\times6},150^{\times2}]$ \\
		CEC2006-g10 & 8  & 6  & $[0^{\times6}]$  & $[5000^{\times3},150000^{\times3}]$ \\
		CEC2010-c13 & 10 & 3  & $[0^{\times3}]$  & $[200^{\times3}]$ \\
		Ackley-2D   & 2  & 2  & $[0^{\times2}]$  & $[200,50]$ \\
		Ackley-6D   & 6  & 2  & $[0^{\times2}]$  & $[200,50]$ \\
		Ackley-10D  & 10 & 2  & $[0^{\times2}]$  & $[200,50]$ \\
		GKXWC1      & 2  & 1  & $[0]$            & $[2]$ \\
		GKXWC2      & 2  & 1  & $[0]$            & $[5]$ \\
		KeaneBump   & 18 & 2  & $[0^{\times2}]$  & $[2,100]$ \\
		PressureVessel & 4 & 4 & $[0^{\times4}]$ & $[10,10,1000,100]$ \\
		RCBeam      & 3  & 2  & $[0^{\times2}]$  & $[10,500]$ \\
		SpeedReducer & 7 & 9  & $[0^{\times9}]$  & $[10^{\times9}]$ \\
		ThreeTruss  & 2  & 3  & $[0^{\times3}]$  & $[10^{\times3}]$ \\
		Car         & 11 & 10 & $[0^{\times10}]$ & $[10^{\times10}]$ \\
		\hline
	\end{tabular}
\end{table}

\section{Additional Implementation Details and Parameter Settings}
\label{app:algorithm_details}

This section provides additional implementation details and parameter settings that are not fully specified in the main paper. These settings are used for all experiments unless otherwise stated.

\subsection{CBA-BO Implementation Details}

All design variables and threshold vectors are normalized into the unit hypercube before model training and candidate generation. The proposed method is implemented in PyTorch and BoTorch using double precision. At each iteration, the objective and each constraint are modeled by independent Gaussian process surrogates. Specifically, we use a ModelListGP, where each output is modeled by a SingleTaskGP. The input variables are normalized, and each output is standardized before GP training. The GP hyperparameters are learned by maximizing the sum of marginal log likelihoods.

The confidence-bound coefficient used in GP-guided learning is set to $\beta=0.5$. Therefore, the objective guidance uses $\mathrm{LCB}_f(\mathbf{x})=\mu_f(\mathbf{x})-0.5\sigma_f(\mathbf{x})$, while constraint feasibility is estimated using $\mathrm{UCB}_{c_j}(\mathbf{x})=\mu_{c_j}(\mathbf{x})+0.5\sigma_{c_j}(\mathbf{x})$. For feasible generated solutions, the PCM is updated according to the gradient of the objective LCB. For infeasible generated solutions, the update direction is computed from the gradients of the violated constraint UCBs. The resulting gradient is normalized before backpropagation.

The PCM is optimized using Adam with an initial learning rate of $10^{-3}$. A cosine annealing learning-rate scheduler is used during training, and the minimum learning rate is set to $10^{-5}$. At each BO iteration, the PCM is updated for 50 gradient steps. The mini-batch size of threshold vectors for each PCM update is 64. The monotonic guidance term is activated in the second half of the training process, and its coefficient is set to 0.1.

\subsection{Parametric Constraint Model}

\paragraph{Model Structure.}
The PCM $h_{\phi}(\boldsymbol{\theta})$ is implemented as a multilayer perceptron (MLP). The input and output dimensions correspond to the number of constraints $m$ and decision variables $D$, respectively. The network contains three hidden layers with 256 neurons and uses SiLU activation functions. The model architecture is

\begin{equation}
	\begin{aligned}
		h_{\phi}(\boldsymbol{\theta})
		:\quad
		\boldsymbol{\theta}
		&\rightarrow
		\text{Linear}(m,256)
		\rightarrow
		\text{SiLU}
		\\
		&\rightarrow
		\text{Linear}(256,256)
		\rightarrow
		\text{SiLU}
		\\
		&\rightarrow
		\text{Linear}(256,256)
		\rightarrow
		\text{SiLU}
		\\
		&\rightarrow
		\text{Linear}(256,D)
		\rightarrow
		\text{Sigmoid}
		\rightarrow
		\mathbf{x}.
	\end{aligned}
\end{equation}

The sigmoid activation in the output layer constrains the predicted solutions within the normalized decision space $[0,1]^D$. The hidden layers are initialized using orthogonal initialization, while the output layer is initialized with small weights to stabilize early-stage optimization.

\subsection{Threshold Batch Sampling}

During PCM training, threshold vectors are sampled by a structured mixture strategy. For each mini-batch, one quarter of the thresholds are uniformly sampled along the main diagonal of the threshold space, one quarter are randomly sampled along the same diagonal, and the remaining half are generated by Sobol sampling over the full threshold domain. This sampling strategy provides both structured coverage of the main threshold relaxation path and global coverage of the whole threshold space.

Let
\begin{equation}
	\Theta=[\boldsymbol{\theta}^{L},\boldsymbol{\theta}^{U}]
	\subset \mathbb{R}^{m},
\end{equation}
where $\boldsymbol{\theta}^{L}$ and $\boldsymbol{\theta}^{U}$ denote the lower and upper bounds of the threshold space. For a mini-batch with size $B = 64$, the sampled threshold set is constructed as
\begin{equation}
	\mathcal{B}_{\theta}
	=
	\mathcal{B}_{d}
	\cup
	\mathcal{B}_{r}
	\cup
	\mathcal{B}_{s},
	\quad
	|\mathcal{B}_{d}|=\frac{B}{4},
	\quad
	|\mathcal{B}_{r}|=\frac{B}{4},
	\quad
	|\mathcal{B}_{s}|=\frac{B}{2}.
\end{equation}
The uniformly sampled diagonal subset is
\begin{equation}
	\mathcal{B}_{d}
	=
	\left\{
	\boldsymbol{\theta}^{L}
	+
	\lambda_i
	(\boldsymbol{\theta}^{U}-\boldsymbol{\theta}^{L})
	\right\}_{i=1}^{B/4},
	\quad
	\lambda_i=
	\frac{i-1}{B/4-1}.
\end{equation}
The randomly sampled diagonal subset is
\begin{equation}
	\mathcal{B}_{r}
	=
	\left\{
	\boldsymbol{\theta}^{L}
	+
	\lambda_i
	(\boldsymbol{\theta}^{U}-\boldsymbol{\theta}^{L})
	\right\}_{i=1}^{B/4},
	\quad
	\lambda_i \sim \mathcal{U}(0,1).
\end{equation}
The global Sobol subset is
\begin{equation}
	\mathcal{B}_{s}
	=
	\left\{
	\boldsymbol{\theta}^{L}
	+
	\mathbf{u}_i \odot
	(\boldsymbol{\theta}^{U}-\boldsymbol{\theta}^{L})
	\right\}_{i=1}^{B/2},
	\quad
	\mathbf{u}_i \sim \mathrm{Sobol}(m),
\end{equation}
where $\odot$ denotes element-wise multiplication.

\subsection{Sampling Strategy}

At each BO iteration, $1000$ threshold vectors are sampled from the threshold domain using the same sampling strategy as that adopted for PCM training, and are subsequently mapped to candidate solutions using the PCM:
\[
\mathcal{X}_t=\{h_{\phi}(\boldsymbol{\theta})\mid
\boldsymbol{\theta}\sim\Theta\}.
\]

To improve the coverage of different constraint requirements, five candidates are selected from the candidate pool. A candidate is considered feasible under a threshold vector $\boldsymbol{\theta}$ if 
$
v_j(\mathbf{x})
=
\max(0,UCB_{c_j}(\mathbf{x})-\theta_j)
=
0,
\quad j=1,\ldots,m .
$ Two candidates are selected according to the extreme threshold settings. For the strict threshold $\boldsymbol{\theta}^{L}$ and the relaxed threshold $\boldsymbol{\theta}^{U}$, the feasible candidate with the lowest $LCB_f(\mathbf{x})$ is selected. If no feasible candidate exists under the corresponding threshold setting, the candidate with the minimum total constraint violation is selected instead.

The remaining three candidates are randomly selected from the intermediate threshold configurations to maintain exploration. A similarity threshold of $0.05$ is imposed in the normalized design space to avoid redundant evaluations. The selected candidates are then evaluated using the expensive objective and constraint functions. 

\subsection{Local Refinement Settings} For the refinement setting, the predicted solution generated by the PCM is first evaluated using the true objective and constraint functions. Then, one additional expensive evaluation is selected by a local BO refinement procedure around the predicted solution. The local search region is constructed in the normalized design space, where all design variables lie in the unit hypercube $[0,1]^D$. The local radius is set to $r=0.25$ along each normalized dimension, and the resulting region is clipped to the global design domain. The local refinement reuses the collected database to fit local GP surrogates and applies LogcEI as the acquisition function. No additional initialization evaluations are used in this stage. After the additional expensive evaluation, the final refined solution is selected from the predicted solution and the locally evaluated candidate using the same feasibility rule as in the main experiments.

\subsection{Constraint-Bound Recommendation Settings}

For the intent-guided constraint-bound recommendation case study, local threshold candidates are sampled around the baseline threshold $\boldsymbol{\theta}^{base}$ in the normalized threshold space.

For each recommendation query, $50000$ local threshold candidates are generated. Among them, $80\%$ are sampled according to the user-specified intents, and $20\%$ are sampled using unbiased Gaussian perturbations for local exploration. For the intent-independent exploration subset, the standard deviation of the perturbation is set to $0.4r_j$ for the $j$-th threshold dimension. For tightening intents, the perturbation $\delta_j^{(i)}$ is sampled toward smaller thresholds using a half-normal distribution with scale $0.5r_j$; for loosening intents, it is sampled toward larger thresholds with the same scale; and for locked constraints, only a very small neighborhood around the baseline is explored, with the perturbation range proportional to $0.05r_j$. All sampled threshold vectors are clipped to the normalized threshold domain $[0,1]^m$.

Each sampled threshold vector $\boldsymbol{\theta}^{(i)}$ is mapped to a candidate solution $\hat{\mathbf{x}}^{*(i)}$ by the learned PCM. The candidate is retained only if the GP mean prediction satisfies all recommended thresholds with a feasibility tolerance $\tau=10^{-4}$. The remaining feasible candidates are ranked by the utility score $S(\boldsymbol{\theta}^{(i)})$ defined in the main text, which combines predicted objective improvement, rewards for successful tightening, and penalties for intent-violating relaxation.

For the utility score, the objective improvement weight is set to $w_f=100$. For tightening and locked intents, relaxation beyond the baseline threshold is penalized with a hard-barrier coefficient $\lambda_h=10000$. For tightening intents, successful tightening is rewarded with $\lambda_r=50$. For loosening intents, the relaxation allowance is defined as $\epsilon_j=\max(0.05|\theta_j^{base}|,1.0)$, and only excessive relaxation beyond this allowance is penalized with a quadratic coefficient of $50$.

\begin{table}[h]
	\centering
	\caption{Main parameter settings used in CBA-BO.}
	\label{tab:parameter_settings}
	\renewcommand{\arraystretch}{1.15}
	\begin{tabular}{lc}
		\hline
		Parameter & Value \\
		\hline
		Initial LHS samples & 30 \\
		Total CBA-BO training evaluations & 200 \\
		BO iterations & 34 \\
		Batch size per BO iteration & 5 \\
		PCM training steps per iteration & 50 \\
		Threshold mini-batch size & 64 \\
		PCM optimizer & Adam \\
		Initial learning rate & $10^{-3}$ \\
		Minimum learning rate & $10^{-5}$ \\
		Learning-rate scheduler & Cosine annealing \\
		LCB/UCB coefficient $\beta$ & 0.5 \\
		Monotonic guidance coefficient $\alpha$ & 0.1 \\
		PCM candidate threshold samples  & 1,000 \\
		Candidate similarity threshold $\delta$ & 0.05 \\
		Local refinement radius $r$ & 0.25 \\
		Refinement evaluations per threshold & 1 \\
		Recommendation local samples  & 50,000 \\
		Recommendation intent-guided ratio & 0.8 \\
		Recommendation exploration ratio & 0.2 \\
		Recommendation objective weight $w_f$ & 100 \\
		Hard-barrier penalty $\lambda_h$ & 10000 \\
		Tightening reward coefficient $\lambda_r$ & 50 \\
		\hline
	\end{tabular}
\end{table}

\section{Additional Experimental Results}

\subsection{Detailed Results on the Full Benchmark Set}

Tables~\ref{tab:detail_direct} and~\ref{tab:detail_refine} provide the complete win-rate results of CBA-BO under the Direct Prediction and Refinement settings, respectively. These tables report the win-rate results for all tested problems, all baseline algorithms, and all numbers of threshold configurations, complementing the summarized results in the main paper.

As shown in Table~\ref{tab:detail_direct}, Direct Prediction alone does not always achieve strong performance. Although the learned PCM can quickly generate a candidate solution for each queried threshold vector, its prediction may still deviate from the true optimum, especially on problems with active constraints, narrow feasible regions, or complex threshold-dependent solution manifolds. Therefore, direct PCM prediction should be viewed as a fast initial recommendation rather than a fully optimized solution.

Table~\ref{tab:detail_refine} shows that one-step refinement consistently improves the results on many problems. By performing one additional local LogcEI evaluation around the PCM prediction, CBA-BO can correct local prediction errors and substantially improve feasibility and solution quality. The improvement from Direct Prediction to Refinement also confirms the role of the proposed two-stage strategy: the PCM provides an efficient global threshold-to-solution prediction, while local refinement improves reliability when the direct prediction is not sufficiently accurate.

To provide a more intuitive view of the optimization results beyond the summarized win-rate statistics, Tables~\ref{tab:single_theta_N1}--\ref{tab:single_theta_N50} present representative objective values for different numbers of threshold configurations. For each $N$, one threshold configuration is selected from the tested cases for presentation due to the large number of evaluated configurations. The tables compare the results obtained by CBA-BO (Direct Prediction and Refinement) with all baseline algorithms on the complete benchmark set. Red entries indicate infeasible solutions, while bold entries denote the best objective values among feasible solutions. The Win Count reports the number of problems where each method achieves the best feasible objective among the selected cases. For PCM, the value in parentheses indicates the cases where the direct prediction already achieves the final best result without refinement.

The detailed results provide additional evidence supporting the win-rate analysis in the main paper. Direct PCM prediction can already obtain competitive solutions for many problems, demonstrating the effectiveness of the learned threshold-solution relationship. Meanwhile, Refinement further improves the predicted solutions in several cases by correcting local prediction errors. These representative results are consistent with the overall win-rate comparison and further illustrate the effectiveness of the proposed two-stage optimization strategy.

\subsection{Visualization of Threshold-to-Solution Prediction}

These visualizations further reveal how the difficulty of threshold-to-solution prediction depends on whether the constraints are active around the optimum. For GKXWC1, the constraint is mostly inactive under the tested threshold settings. Therefore, the optimum is mainly determined by the objective landscape rather than by the constraint boundary. As a result, the optimal solutions concentrate in a small region, and the threshold-to-solution mapping is relatively simple. This explains why the PCM predictions almost overlap with the ground-truth optima on GKXWC1.

When constraints become active, the optimum is pushed onto or along the feasible boundary, and the solution distribution becomes more dependent on the threshold values. GKXWC2 illustrates this case. As the threshold changes, the feasible boundary moves, and the optimal solutions form a narrow boundary-like distribution rather than a single compact cluster. Nevertheless, the mapping remains relatively smooth and mostly single-valued, so the PCM can still accurately track the movement of the optimum.

The prediction becomes more difficult when the active constraint structure changes across thresholds. ThreeTruss shows this behavior: the ground-truth optima are more dispersed and form a curved, non-uniform solution manifold. Different threshold settings may activate different constraints, causing the optimum to move between different boundary segments. In such cases, the threshold-to-solution mapping is less smooth and harder to approximate globally. The PCM can still capture the overall distribution trend, but local prediction errors are more likely to occur.

Overall, direct PCM prediction is most reliable when the constraints are inactive or when the active constraints induce a smooth and continuous movement of the optimum. It becomes more challenging when the feasible region is narrow, disconnected, or when small threshold changes lead to changes in the active constraint set. These observations explain the role of the local LogcEI refinement: the PCM provides fast global prediction over the threshold space, while refinement corrects difficult cases caused by active or changing constraint boundaries.

\subsection{Analysis of PCM-based Initialization for Refinement}

To further analyze the contribution of the learned threshold-solution relationship, we compare the refinement results of CBA-BO with a random-initialization LogcEI baseline. The purpose of this experiment is to investigate whether the performance improvement mainly comes from the PCM-guided initialization rather than the LogcEI refinement itself. In CBA-BO, the refinement step is mainly used to correct the approximation errors of the PCM prediction, including deviations from the optimal solution and possible infeasible fluctuations, thereby improving the quality of the predicted solutions.

\begin{table}[t]
	\centering
	\caption{Win rates of PCM-guided refinement against random-initialization refinement over 1,000 threshold configurations.}
	\label{tab:pcm_refinement}
	\begin{tabular}{lc}
		\hline
		Problem &  Win Rate (\%) \\
		\hline
		CEC2006-g01 & 47.5 \\
		CEC2006-g06 & 98.4 \\
		CEC2006-g07 & 89.3 \\
		CEC2006-g10 & 99.9 \\
		CEC2010-c13 & 99.9 \\
		Ackley-10D & 98.6 \\
		Ackley-2D & 100.0 \\
		Ackley-6D & 100.0 \\
		Car & 92.6 \\
		GKXWC1 & 82.9 \\
		GKXWC2 & 45.5 \\
		KeaneBump & 97.9 \\
		PressureVessel & 97.6 \\
		ReinforcedConcreteBeam & 99.1 \\
		SpeedReducer & 59.1 \\
		ThreeTruss & 100.0 \\
		\hline
	\end{tabular}
\end{table}

The CBA-BO results are directly obtained from the one-step refinement procedure described in the main paper. For comparison, we construct a baseline by randomly sampling 200 solutions from the design space and training the GP surrogate models using these samples. Given each target threshold configuration, the LogcEI acquisition function is optimized once to obtain the refined solution. The refinement procedure and acquisition function are identical to those used in CBA-BO, while the only difference is the source of the initial solution and surrogate training data.

The comparison is performed over 1,000 randomly sampled threshold configurations for each problem. Table~\ref{tab:pcm_refinement} reports the win rates of CBA-BO refinement against random-initialization LogcEI refinement. CBA-BO achieves a high win rate across all problems, with particularly strong performance on CEC2006-g10, CEC2010-c13, Ackley-2D, Ackley-6D, KeaneBump, PressureVessel, RCBeam, and ThreeTruss. These results demonstrate that the learned threshold-solution relationship provides valuable guidance for refinement, allowing CBA-BO to start from more informative solutions and effectively correct prediction errors during refinement.

For several problems, such as CEC2006-g01 and GKXWC2, the random-initialization baseline remains competitive. This is mainly because these problems have relatively simple objective and constraint landscapes, where the GP surrogate trained with 200 randomly sampled solutions can already provide sufficiently accurate approximations. Therefore, the advantage of PCM-based initialization becomes less significant in such cases. Overall, the results confirm that the performance improvement of CBA-BO is not solely attributed to the final LogcEI refinement, but also benefits from the proposed PCM-based threshold-aware learning strategy. The refinement step further improves the predicted solutions by correcting approximation errors and infeasible fluctuations introduced by the learned mapping.

\subsection {Monotonicity Analysis of Monotonic Regularization}

To quantitatively evaluate the effect of monotonic regularization on the learned threshold-response relationship, we further conduct a pairwise monotonicity test. Specifically, we randomly sample 10,000 pairs of constraint thresholds $(\boldsymbol \theta_1,\boldsymbol \theta_2)$ satisfying $\boldsymbol \theta_1<\boldsymbol \theta_2$ and evaluate whether the corresponding predicted objective values satisfy the expected monotonic relationship, i.e., $f(x_2)\leq f(x_1)$. The monotonicity rate is calculated as the percentage of pairs satisfying this condition. The results are summarized in Table~\ref{tab:monotonicity}.

The results show that monotonic regularization generally improves the monotonic consistency of the learned threshold-response relationship. Significant improvements are observed on several challenging problems, including SpeedReducer (+59.7\%), KeaneBump (+32.0\%), RCBeam (+22.9\%), and GKXWC2 (+20.8\%). These results indicate that the introduced monotonic prior effectively suppresses irregular variations in the parametric constraint model when the underlying threshold-solution relationship is difficult to learn. Similar improvements can also be observed on CEC2006-g06 and CEC2006-g10, where the monotonicity rate reaches 100.0\% after regularization.

Meanwhile, monotonic regularization does not introduce noticeable degradation on problems where the original model already captures a naturally monotonic relationship. For PressureVessel, ThreeTruss, and Car, both models achieve a monotonicity rate of 100.0\%, demonstrating that the regularization preserves the existing monotonic structure without imposing unnecessary bias.

Although a slight decrease is observed on CEC2006-g07, the overall monotonicity remains high. This suggests that monotonic regularization is not designed to enforce monotonicity at the expense of model flexibility, but rather provides a structural preference that improves response consistency in problems where monotonic violations are more prominent.

\begin{table}[t]
	\centering
	\caption{Objective-only pairwise monotonicity test ($N=10000$ pairs). For each pair $(\boldsymbol\theta_1, \boldsymbol\theta_2)$ with $\boldsymbol\theta_1 < \boldsymbol\theta_2$, we check if $f(x_2) \leq f(x_1)$.}
	\label{tab:monotonicity}
	\renewcommand{\arraystretch}{1.3}
	\setlength{\tabcolsep}{2pt}
	\begin{tabular}{@{}lcccc@{}}
		\hline
		Problem & No Mono (\%) & With Mono (\%) & $\Delta$ (\%) \\
		\hline
		CEC2006-g01 & 98.7 & \textbf{99.7} & +1.0 \\
		CEC2006-g06 & 85.1 & \textbf{100.0} & +14.9 \\
		CEC2006-g07 & \textbf{81.9} & 72.5 & -9.4 \\
		CEC2006-g10 & 87.2 & \textbf{100.0} & +12.8 \\
		CEC2010-c13 & 82.5 & \textbf{86.0} & +3.5 \\
		GKXWC1 & 44.8 & \textbf{59.5} & +14.7 \\
		GKXWC2 & 75.5 & \textbf{96.3} & +20.8 \\
		KeaneBump & 61.7 & \textbf{93.8} & +32.0 \\
		RCBeam & 77.1 & \textbf{100.0} & +22.9 \\
		SpeedReducer & 40.3 & \textbf{100.0} & +59.7 \\
		PressureVessel & 100.0 & 100.0 & 0.0 \\
		ThreeTruss & 100.0 & 100.0 & 0.0 \\
		Car & 100.0 & 100.0 & 0.0 \\
		\hline
	\end{tabular}
\end{table}

\subsection {Run time}

We report the wall-clock time of one optimization iteration for each method in Table~\ref{tab:runtime}. For CBA-BO, we additionally provide the detailed computational cost of PCM training, PCM prediction, and one-step refinement. The PCM training time represents the cost of updating the threshold-solution relationship based on the current surrogate models, while the PCM prediction time measures the overhead of generating solutions for new threshold configurations. The results show that PCM prediction introduces negligible computational overhead, requiring only several milliseconds for all problems. The refinement step incurs additional computational cost due to the optimization of the acquisition function, but it effectively corrects prediction errors and infeasible fluctuations of PCM-generated solutions, thereby improving the final solution quality.

For conventional constrained BO methods, a given threshold configuration is usually optimized independently through a sequential optimization process, requiring tens or hundreds of expensive evaluations before obtaining a satisfactory solution. When different threshold configurations need to be explored, this optimization process must be repeatedly performed from scratch. In contrast, after the PCM is learned within the optimization process (34 iterations in our experiments), CBA-BO can directly predict solutions for unseen threshold configurations with negligible prediction overhead and further improve them through refinement.

In expensive constrained optimization problems, the dominant computational cost usually comes from objective and constraint evaluations, which may involve time-consuming simulations or real-world experiments. The internal computational overhead of all optimization algorithms, including surrogate modeling and acquisition optimization, is generally negligible compared with the expensive evaluations. Therefore, the number of optimization iterations, which directly determines the number of expensive evaluations, is a more critical factor for evaluating optimization efficiency. By learning the threshold-solution relationship, CBA-BO avoids repeatedly performing long optimization processes for different threshold configurations and provides an efficient solution for continuously varying constraint thresholds.

\begin{table*}[t]
	\centering
	\caption{Wall-clock time (seconds) of one optimization iteration. For CBA-BO, the computational costs of PCM training, PCM prediction, and refinement are separately reported.}
	\label{tab:runtime}
	\renewcommand{\arraystretch}{1.2}
	\setlength{\tabcolsep}{4pt}
	\begin{tabular}{lccccccccc}
		\hline
		Problem & PCM Training & PCM Prediction & Refinement & cEI & LogcEI & ALBO & SCBO & CMA-ES \\
		\hline
		CEC2006-g01 & 11.970 & 0.0003 & 5.318 & 5.264 & 9.931 & 34.990 & 3.458 & 0.0055 \\
		CEC2006-g06 & 7.890 & 0.0002 & 4.476 & 10.967 & 11.168 & 11.814 & 5.986 & 0.0010 \\
		CEC2006-g07 & 20.173 & 0.0002 & 16.435 & 12.241 & 14.767 & 26.281 & 11.644 & 0.0020 \\
		CEC2006-g10 & 12.885 & 0.0003 & 8.881 & 6.570 & 8.353 & 47.601 & 6.602 & 0.0010 \\
		CEC2010-c13 & 8.590 & 0.0002 & 6.396 & 1.792 & 2.794 & 21.075 & 2.124 & 0.0010 \\
		Ackley-10D & 5.447 & 0.0003 & 1.586 & 1.178 & 1.532 & 3.298 & 0.935 & 0.0010 \\
		Ackley-2D & 2.224 & 0.0002 & 1.100 & 0.573 & 0.630 & 1.547 & 0.361 & 0.0010 \\
		Ackley-6D & 4.155 & 0.0001 & 2.566 & 1.049 & 1.451 & 2.123 & 0.770 & 0.0010 \\
		GKXWC1 & 1.085 & 0.0004 & 0.442 & 0.534 & 0.573 & 1.407 & 0.440 & 0.0010 \\
		GKXWC2 & 1.103 & 0.0002 & 0.489 & 0.255 & 0.379 & 0.847 & 0.200 & 0.0010 \\
		KeaneBump & 16.019 & 0.0002 & 7.608 & 4.847 & 5.192 & 14.768 & 4.527 & 0.0045 \\
		PressureVessel & 5.935 & 0.0001 & 3.044 & 4.003 & 2.937 & 32.946 & 2.413 & 0.0010 \\
		ReinforcedConcreteBeam & 1.996 & 0.0001 & 0.639 & 0.742 & 0.976 & 1.097 & 1.085 & 0.0010 \\
		SpeedReducer & 17.971 & 0.0002 & 12.108 & 15.340 & 16.535 & 15.524 & 10.928 & 0.0020 \\
		ThreeTruss & 2.867 & 0.0004 & 1.678 & 1.122 & 1.631 & 3.613 & 0.822 & 0.0010 \\
		\hline
	\end{tabular}
\end{table*}

\section{Parameter Analysis}
\label{app:parameter_analysis}

We further analyze two important hyperparameters of CBA-BO: the confidence-bound coefficient used in GP-guided learning and the number of inner PCM training steps. The experiments are conducted on four representative problems, including CEC2006-g10, CEC2010-c13, PressureVessel, and ThreeTruss. For each alternative parameter setting, we compare it with the default setting used in the main experiments. Both settings are independently run five times, and the win rate (\%) of the default setting against the alternative setting is reported. A value larger than $50\%$ indicates that the default setting performs better on more threshold configurations.

\subsection{Effect of the Confidence-Bound Coefficient}

The confidence-bound coefficient $\beta$ controls the influence of GP uncertainty in the LCB/UCB-based guidance. A small $\beta$ relies more on the surrogate mean prediction, while a larger $\beta$ introduces stronger uncertainty-aware guidance. The default setting uses $\beta=0.5$.

\begin{table}[h]
	\centering
	\caption{Win rate (\%) of the default setting against alternative confidence-bound coefficients $\beta$.}
	\label{tab:lcb_analysis}
	\small
	\renewcommand{\arraystretch}{1.05}
	\setlength{\tabcolsep}{4pt}
	\begin{tabular}{lcccc}
		\hline
		Alternative setting & g10 & c13 & PressureVessel & ThreeTruss \\
		\hline
		$\beta=0.01$ & 70.3 & 43.0 & 58.1 & {73.0} \\
		$\beta=1.0$  & {75.7} & 52.1 & 69.1 & 71.5 \\
		$\beta=2.0$  & 69.5 & {54.9} & {76.7} & 72.4 \\
		\hline
	\end{tabular}
\end{table}

As shown in Table~\ref{tab:lcb_analysis}, the default setting obtains win rates above $50\%$ in most cases, indicating that it provides a stable balance between surrogate mean exploitation and uncertainty-aware guidance. The only exception is CEC2010-c13 with $\beta=0.01$, where the default setting achieves a win rate of $43.0\%$, suggesting that a smaller uncertainty coefficient can be competitive on this problem. However, on PressureVessel and ThreeTruss, the default setting consistently outperforms all alternative $\beta$ values. These results show that the chosen confidence-bound coefficient is generally robust across different problems.

\subsection{Effect of the Number of Inner PCM Training Steps}

We also investigate the number of inner PCM training steps performed
after each GP fitting stage. This parameter controls how many gradient
updates are applied to the PCM within each BO iteration. The default setting uses 50 inner PCM training steps.

\begin{table}[h]
	\centering
	\caption{Win rate (\%) of the default setting against alternative numbers of inner PCM training steps.}
	\label{tab:inner_analysis}
	\small
	\renewcommand{\arraystretch}{1.05}
	\setlength{\tabcolsep}{4pt}
	\begin{tabular}{lcccc}
		\hline
		Alternative setting & g10 & c13 & PressureVessel & ThreeTruss \\
		\hline
		Inner $=20$  & {99.1} & 50.7 & {57.5} & {78.4} \\
		Inner $=100$ & 20.5 & 47.6 & 44.6 & 48.8 \\
		Inner $=200$ & 0.2 & {55.5} & 35.5 & 48.6 \\
		\hline
	\end{tabular}
\end{table}

Table~\ref{tab:inner_analysis} shows that the effect of the number of inner PCM training steps is problem-dependent. The default setting outperforms Inner $=20$ on PressureVessel and ThreeTruss, suggesting that too few PCM updates may be insufficient for learning the threshold-to-solution mapping on problems where constraints strongly affect the optimum. On CEC2006-g10, however, the constraints are inactive in most tested threshold settings. Therefore, the comparison is mainly determined by the numerical accuracy of the predicted objective value rather than by constraint satisfaction. Since the final feasible solutions are very similar in this case, using more inner updates can lead to slightly more accurate objective optimization, which explains the low win rates of the default setting against Inner $=100$ and Inner $=200$ on CEC2006-g10. Nevertheless, increasing the number of inner steps does not consistently improve performance on the other problems and may increase the risk of over-adapting the PCM to the current GP surrogate. Therefore, the default setting adopts a moderate number of inner updates to maintain a stable trade-off between learning effectiveness, robustness, and computational efficiency.

\subsection{Effect of the Monotonic Regularization Coefficient}

The monotonic regularization coefficient $\alpha$ controls the strength of the monotonic constraint imposed on PCM. A smaller $\alpha$ provides a weaker monotonic regularization, while a larger $\alpha$ enforces stronger monotonic guidance during model learning. The default setting uses
$\alpha=0.1$.

\begin{table}[h]
	\centering
	\caption{Win rate (\%) of the default setting against alternative monotonic regularization coefficients $\alpha$.}
	\label{tab:alpha_analysis}
	\small
	\renewcommand{\arraystretch}{1.05}
	\setlength{\tabcolsep}{4pt}
	\begin{tabular}{lcccc}
		\hline
		Alternative setting & g10 & c13 & PressureVessel & ThreeTruss \\
		\hline
		$\alpha=0.01$ & 100.0 & 55.1 & 77.5 & 46.2 \\
		$\alpha=0.5$  & 100.0 & 50.0 & 68.6 & 49.7 \\
		$\alpha=1.0$  & 100.0 & 63.4 & 89.8 & 42.1 \\
		\hline
	\end{tabular}
\end{table}

As shown in Table~\ref{tab:alpha_analysis}, the default setting achieves competitive performance against different monotonic regularization coefficients. The results indicate that PCM is robust to the choice of $\alpha$, with the default setting outperforming alternative coefficients in most cases. Although different problems exhibit different preferences for the regularization strength, no consistent advantage is observed for any alternative setting. This suggests that the effectiveness of PCM mainly comes from exploiting the underlying monotonic relationship between constraint thresholds and solutions, rather than relying on a specific regularization coefficient.

\begin{figure*}[t]
	\centering
	\includegraphics[width=0.8\textwidth]{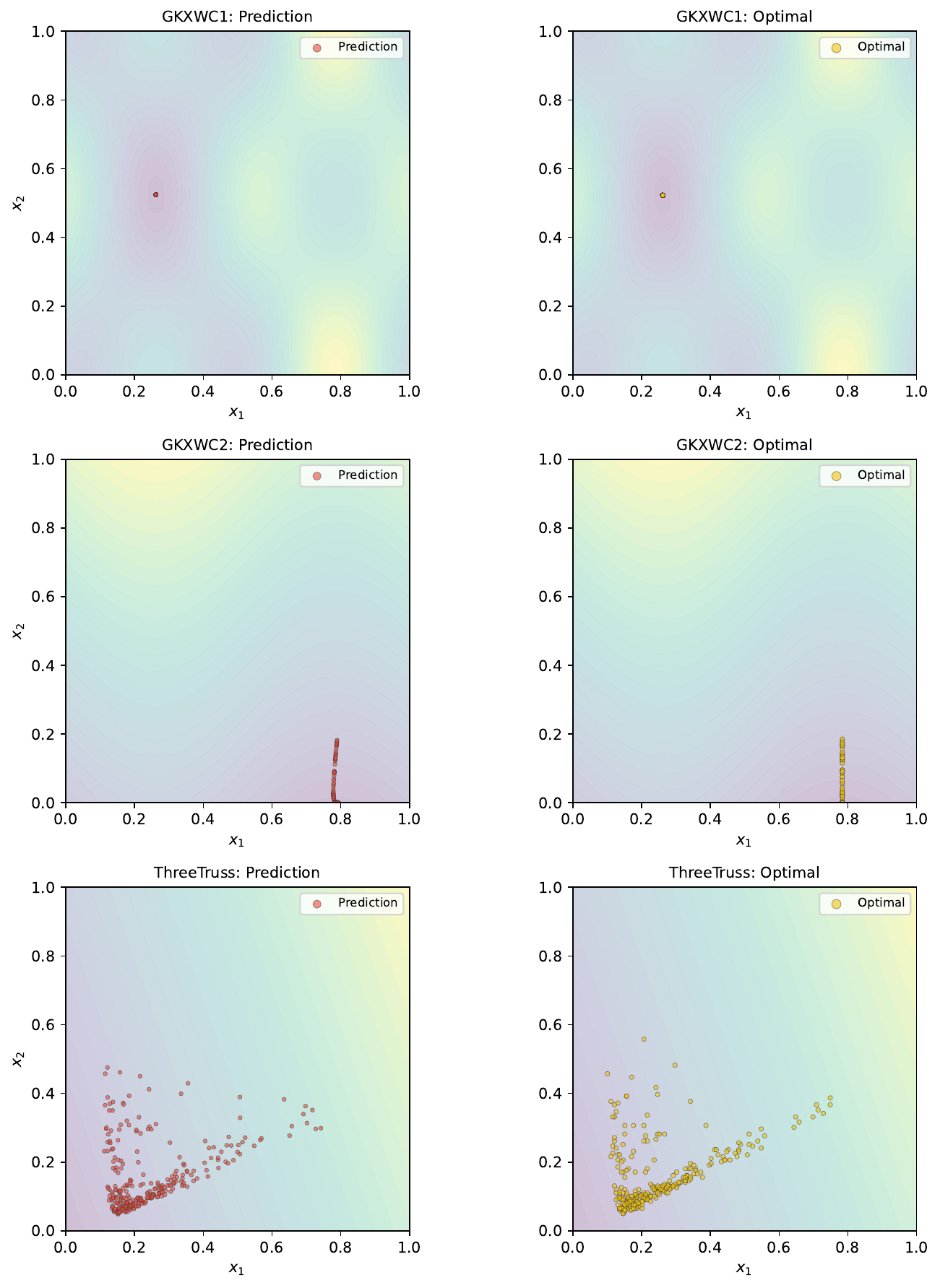}
	\caption{
		Distribution of PCM-predicted solutions and grid-search optima on three 2D problems.
		For each problem, the left panel shows PCM predictions and the right panel shows the corresponding grid-search optima over 300 randomly sampled threshold settings.
	}
	\label{fig:distribution}
\end{figure*}

\begin{figure*}[t]
	\centering
	\includegraphics[width=\textwidth]{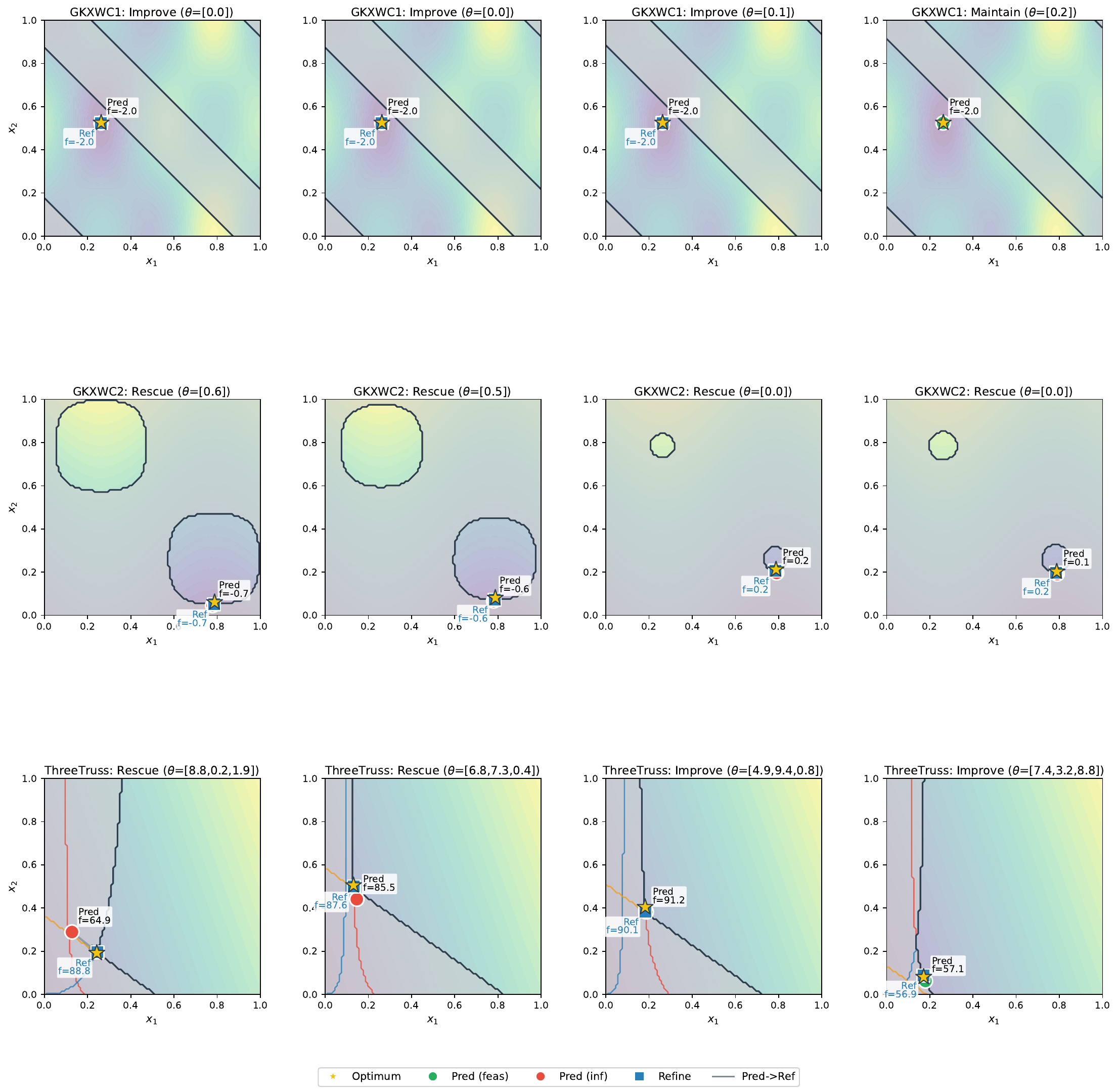}
	\caption{
		Visualization of PCM prediction and LogcEI refinement on three 2D problems.
		Gold stars denote grid-search optima, green circles denote feasible PCM predictions, red circles denote infeasible PCM predictions, and blue squares denote refined solutions. Arrows connect each PCM prediction to its refined solution.
	}
	\label{fig:showcase}
\end{figure*}

\begin{table*}[t]
	\centering
	\caption{Detailed win rates (\%) of direct prediction against baseline algorithms. Each block reports results under $N\in\{1,5,10,20,50\}$ threshold configurations. }
	\label{tab:detail_direct}
	\renewcommand{\arraystretch}{1.3}
	\setlength{\tabcolsep}{2pt}
	\begin{minipage}{0.5\textwidth}
		\centering
		\begin{tabular}{@{}llccccc@{}}
			\hline
			Problem & $N$ & CMA-ES & cEI & ALBO & SCBO & LogcEI \\
			\hline
			CEC2006-g01 & 1 & 20.0 & 20.0 & 10.0 & 0.0 & 0.0 \\
			& 5 & 36.0 & 38.0 & 32.0 & 2.0 & 0.0 \\
			& 10 & 37.0 & 31.0 & 39.0 & 2.0 & 0.0 \\
			& 20 & 41.5 & 49.0 & 49.0 & 4.5 & 0.0 \\
			& 50 & 42.6 & 54.8 & 56.0 & 7.8 & 3.6 \\
			\hline
			CEC2006-g04 & 1 & 90.0 & 80.0 & 80.0 & 90.0 & 80.0 \\
			& 5 & 100.0 & 100.0 & 100.0 & 100.0 & 100.0 \\
			& 10 & 97.0 & 95.0 & 95.0 & 97.0 & 95.0 \\
			& 20 & 98.5 & 98.0 & 98.0 & 98.5 & 98.0 \\
			& 50 & 98.4 & 98.2 & 97.8 & 98.4 & 97.4 \\
			\hline
			CEC2006-g07 & 1 & 70.0 & 70.0 & 80.0 & 0.0 & 0.0 \\
			& 5 & 78.0 & 74.0 & 78.0 & 68.0 & 38.0 \\
			& 10 & 84.0 & 82.0 & 89.0 & 76.0 & 67.0 \\
			& 20 & 78.0 & 74.0 & 82.5 & 59.0 & 59.0 \\
			& 50 & 80.8 & 84.4 & 87.2 & 63.6 & 63.6 \\
			\hline
			CEC2006-g10 & 1 & 100.0 & 100.0 & 90.0 & 100.0 & 80.0 \\
			& 5 & 100.0 & 96.0 & 90.0 & 100.0 & 72.0 \\
			& 10 & 99.0 & 92.0 & 91.0 & 99.0 & 64.0 \\
			& 20 & 100.0 & 95.0 & 95.0 & 100.0 & 67.0 \\
			& 50 & 100.0 & 96.4 & 97.2 & 100.0 & 67.0 \\
			\hline
			CEC2010-c13 & 1 & 100.0 & 70.0 & 80.0 & 0.0 & 50.0 \\
			& 5 & 100.0 & 80.0 & 86.0 & 56.0 & 62.0 \\
			& 10 & 100.0 & 88.0 & 94.0 & 88.0 & 87.0 \\
			& 20 & 98.5 & 97.0 & 95.0 & 95.0 & 95.5 \\
			& 50 & 99.0 & 99.6 & 100.0 & 98.8 & 99.6 \\
			\hline
			Ackley-2D & 1 & 10.0 & 90.0 & 90.0 & 0.0 & 90.0 \\
			& 5 & 90.0 & 90.0 & 90.0 & 76.0 & 90.0 \\
			& 10 & 93.0 & 89.0 & 90.0 & 85.0 & 89.0 \\
			& 20 & 95.0 & 91.5 & 90.0 & 89.0 & 93.5 \\
			& 50 & 97.4 & 91.4 & 90.8 & 89.8 & 100.0 \\
			\hline
			Ackley-6D & 1 & 70.0 & 90.0 & 100.0 & 0.0 & 100.0 \\
			& 5 & 100.0 & 100.0 & 100.0 & 84.0 & 96.0 \\
			& 10 & 100.0 & 99.0 & 100.0 & 100.0 & 100.0 \\
			& 20 & 100.0 & 100.0 & 100.0 & 100.0 & 100.0 \\
			& 50 & 99.8 & 99.8 & 99.8 & 99.6 & 99.8 \\
			\hline
			Ackley-10D & 1 & 60.0 & 60.0 & 100.0 & 10.0 & 80.0 \\
			& 5 & 100.0 & 94.0 & 98.0 & 92.0 & 78.0 \\
			& 10 & 99.0 & 97.0 & 99.0 & 99.0 & 97.0 \\
			& 20 & 100.0 & 100.0 & 100.0 & 100.0 & 100.0 \\
			& 50 & 99.8 & 99.8 & 99.8 & 99.8 & 99.8 \\
			\hline
		\end{tabular}
	\end{minipage}
	\hfill
	\begin{minipage}{0.49\textwidth}
		\centering
		\begin{tabular}{@{}llccccc@{}}
			\hline
			Problem & $N$ & CMA-ES & cEI & ALBO & SCBO & LogcEI \\
			\hline
			GKXWC1 & 1 & 0.0 & 10.0 & 0.0 & 0.0 & 0.0 \\
			& 5 & 86.0 & 40.0 & 48.0 & 16.0 & 6.0 \\
			& 10 & 90.0 & 56.0 & 68.0 & 60.0 & 14.0 \\
			& 20 & 90.0 & 76.0 & 82.5 & 86.5 & 50.0 \\
			& 50 & 90.0 & 86.6 & 87.6 & 89.6 & 86.4 \\
			\hline
			GKXWC2 & 1 & 0.0 & 10.0 & 20.0 & 30.0 & 10.0 \\
			& 5 & 68.0 & 42.0 & 56.0 & 54.0 & 8.0 \\
			& 10 & 83.0 & 59.0 & 61.0 & 72.0 & 19.0 \\
			& 20 & 90.0 & 62.0 & 60.0 & 75.0 & 16.5 \\
			& 50 & 85.8 & 67.0 & 67.2 & 77.2 & 27.0 \\
			\hline
			KeaneBump & 1 & 40.0 & 0.0 & 20.0 & 10.0 & 0.0 \\
			& 5 & 62.0 & 34.0 & 34.0 & 60.0 & 38.0 \\
			& 10 & 56.0 & 47.0 & 43.0 & 66.0 & 43.0 \\
			& 20 & 62.0 & 54.5 & 46.5 & 69.5 & 54.0 \\
			& 50 & 62.0 & 65.4 & 56.4 & 71.6 & 61.8 \\
			\hline
			PressureVessel & 1 & 80.0 & 30.0 & 60.0 & 20.0 & 0.0 \\
			& 5 & 86.0 & 52.0 & 60.0 & 76.0 & 28.0 \\
			& 10 & 84.0 & 55.0 & 69.0 & 83.0 & 46.0 \\
			& 20 & 86.5 & 69.5 & 82.5 & 86.5 & 70.5 \\
			& 50 & 87.0 & 84.8 & 85.2 & 87.0 & 85.0 \\
			\hline
			RCBeam & 1 & 30.0 & 10.0 & 20.0 & 20.0 & 0.0 \\
			& 5 & 82.0 & 10.0 & 22.0 & 48.0 & 0.0 \\
			& 10 & 74.0 & 13.0 & 25.0 & 46.0 & 0.0 \\
			& 20 & 77.0 & 14.5 & 33.0 & 68.5 & 1.0 \\
			& 50 & 79.0 & 29.4 & 48.2 & 71.4 & 2.0 \\
			\hline
			SpeedReducer & 1 & 100.0 & 80.0 & 90.0 & 90.0 & 80.0 \\
			& 5 & 92.0 & 72.0 & 80.0 & 92.0 & 70.0 \\
			& 10 & 88.0 & 68.0 & 78.0 & 88.0 & 68.0 \\
			& 20 & 91.0 & 73.5 & 82.5 & 91.0 & 70.5 \\
			& 50 & 90.4 & 75.2 & 82.2 & 90.4 & 71.4 \\
			\hline
			ThreeTruss & 1 & 70.0 & 40.0 & 50.0 & 0.0 & 40.0 \\
			& 5 & 58.0 & 40.0 & 52.0 & 6.0 & 44.0 \\
			& 10 & 60.0 & 56.0 & 62.0 & 38.0 & 56.0 \\
			& 20 & 63.0 & 60.5 & 60.5 & 53.5 & 60.0 \\
			& 50 & 60.0 & 60.0 & 60.4 & 57.0 & 59.8 \\
			\hline
			Car & 1 & 90.0 & 40.0 & 20.0 & 30.0 & 0.0 \\
			& 5 & 64.0 & 46.0 & 20.0 & 64.0 & 2.0 \\
			& 10 & 74.0 & 32.0 & 13.0 & 74.0 & 2.0 \\
			& 20 & 74.5 & 45.5 & 25.0 & 74.5 & 8.0 \\
			& 50 & 73.6 & 44.0 & 33.8 & 73.6 & 13.6 \\
			\hline
		\end{tabular}
	\end{minipage}
\end{table*}

\begin{table*}[t]
	\centering
	\caption{Detailed win rates (\%) of CBA-BO with one-step refinement against baseline algorithms. Each block reports results under $N\in\{1,5,10,20,50\}$ threshold configurations. }
	\label{tab:detail_refine}
	\renewcommand{\arraystretch}{1.3}
	\setlength{\tabcolsep}{2pt}
	\begin{minipage}{0.5\textwidth}
		\centering
		\begin{tabular}{@{}llccccc@{}}
			\hline
			Problem & $N$ & CMA-ES & cEI & ALBO & SCBO & LogcEI \\
			\hline
			CEC2006-g01 & 1 & 80.0 & 50.0 & 40.0 & 60.0 & 0.0 \\
			& 5 & 98.0 & 88.0 & 66.0 & 92.0 & 0.0 \\
			& 10 & 99.0 & 88.0 & 75.0 & 96.0 & 6.0 \\
			& 20 & 99.5 & 88.5 & 80.5 & 97.0 & 12.0 \\
			& 50 & 98.4 & 90.2 & 85.4 & 95.0 & 46.0 \\
			\hline
			CEC2006-g04 & 1 & 100.0 & 80.0 & 90.0 & 90.0 & 80.0 \\
			& 5 & 100.0 & 100.0 & 100.0 & 100.0 & 100.0 \\
			& 10 & 100.0 & 99.0 & 100.0 & 99.0 & 97.0 \\
			& 20 & 99.5 & 99.5 & 99.5 & 99.5 & 98.5 \\
			& 50 & 99.6 & 99.4 & 99.6 & 99.4 & 98.8 \\
			\hline
			CEC2006-g07 & 1 & 100.0 & 100.0 & 100.0 & 0.0 & 10.0 \\
			& 5 & 100.0 & 100.0 & 100.0 & 100.0 & 64.0 \\
			& 10 & 100.0 & 100.0 & 100.0 & 100.0 & 92.0 \\
			& 20 & 100.0 & 100.0 & 100.0 & 100.0 & 99.5 \\
			& 50 & 100.0 & 100.0 & 100.0 & 100.0 & 100.0 \\
			\hline
			CEC2006-g10 & 1 & 100.0 & 100.0 & 100.0 & 100.0 & 100.0 \\
			& 5 & 100.0 & 100.0 & 100.0 & 100.0 & 100.0 \\
			& 10 & 100.0 & 100.0 & 100.0 & 100.0 & 100.0 \\
			& 20 & 100.0 & 100.0 & 100.0 & 100.0 & 100.0 \\
			& 50 & 100.0 & 100.0 & 100.0 & 100.0 & 99.6 \\
			\hline
			CEC2010-c13 & 1 & 100.0 & 80.0 & 100.0 & 0.0 & 50.0 \\
			& 5 & 100.0 & 80.0 & 98.0 & 70.0 & 70.0 \\
			& 10 & 100.0 & 91.0 & 99.0 & 94.0 & 94.0 \\
			& 20 & 99.0 & 98.5 & 98.0 & 98.5 & 98.5 \\
			& 50 & 99.4 & 99.8 & 100.0 & 99.4 & 99.6 \\
			\hline
			Ackley-2D & 1 & 10.0 & 90.0 & 90.0 & 0.0 & 90.0 \\
			& 5 & 90.0 & 90.0 & 90.0 & 76.0 & 90.0 \\
			& 10 & 93.0 & 89.0 & 90.0 & 85.0 & 89.0 \\
			& 20 & 95.0 & 91.5 & 90.0 & 89.0 & 93.5 \\
			& 50 & 97.4 & 91.4 & 90.8 & 89.8 & 100.0 \\
			\hline
			Ackley-6D & 1 & 70.0 & 90.0 & 100.0 & 0.0 & 100.0 \\
			& 5 & 100.0 & 100.0 & 100.0 & 84.0 & 96.0 \\
			& 10 & 100.0 & 99.0 & 100.0 & 100.0 & 100.0 \\
			& 20 & 100.0 & 100.0 & 100.0 & 100.0 & 100.0 \\
			& 50 & 100.0 & 100.0 & 100.0 & 99.8 & 100.0 \\
			\hline
			Ackley-10D & 1 & 60.0 & 60.0 & 100.0 & 10.0 & 80.0 \\
			& 5 & 100.0 & 94.0 & 98.0 & 92.0 & 78.0 \\
			& 10 & 100.0 & 97.0 & 100.0 & 100.0 & 98.0 \\
			& 20 & 100.0 & 100.0 & 100.0 & 100.0 & 100.0 \\
			& 50 & 100.0 & 99.8 & 100.0 & 100.0 & 99.8 \\
			\hline
		\end{tabular}
	\end{minipage}
	\hfill
	\begin{minipage}{0.49\textwidth}
		\centering
		\begin{tabular}{@{}llccccc@{}}
			\hline
			Problem & $N$ & CMA-ES & cEI & ALBO & SCBO & LogcEI \\
			\hline
			GKXWC1 & 1 & 0.0 & 10.0 & 0.0 & 0.0 & 0.0 \\
			& 5 & 88.0 & 42.0 & 50.0 & 18.0 & 8.0 \\
			& 10 & 90.0 & 60.0 & 69.0 & 62.0 & 19.0 \\
			& 20 & 90.0 & 76.5 & 82.5 & 86.5 & 53.0 \\
			& 50 & 90.0 & 86.8 & 87.6 & 89.8 & 86.4 \\
			\hline
			GKXWC2 & 1 & 30.0 & 50.0 & 50.0 & 60.0 & 30.0 \\
			& 5 & 86.0 & 60.0 & 78.0 & 72.0 & 34.0 \\
			& 10 & 90.0 & 74.0 & 76.0 & 84.0 & 35.0 \\
			& 20 & 96.0 & 77.0 & 80.0 & 90.5 & 44.5 \\
			& 50 & 93.0 & 80.0 & 85.6 & 89.8 & 59.0 \\
			\hline
			KeaneBump & 1 & 50.0 & 0.0 & 20.0 & 10.0 & 0.0 \\
			& 5 & 64.0 & 36.0 & 38.0 & 64.0 & 38.0 \\
			& 10 & 57.0 & 48.0 & 43.0 & 71.0 & 44.0 \\
			& 20 & 63.5 & 56.0 & 48.0 & 71.0 & 54.5 \\
			& 50 & 64.4 & 66.6 & 58.2 & 75.6 & 63.2 \\
			\hline
			PressureVessel & 1 & 90.0 & 30.0 & 60.0 & 20.0 & 0.0 \\
			& 5 & 96.0 & 54.0 & 70.0 & 82.0 & 30.0 \\
			& 10 & 98.0 & 64.0 & 84.0 & 91.0 & 49.0 \\
			& 20 & 98.0 & 78.5 & 93.0 & 95.0 & 77.0 \\
			& 50 & 98.4 & 93.2 & 97.0 & 97.8 & 93.4 \\
			\hline
			RCBeam & 1 & 100.0 & 100.0 & 100.0 & 70.0 & 50.0 \\
			& 5 & 100.0 & 100.0 & 98.0 & 80.0 & 66.0 \\
			& 10 & 100.0 & 100.0 & 100.0 & 97.0 & 56.0 \\
			& 20 & 100.0 & 98.5 & 97.0 & 98.5 & 71.0 \\
			& 50 & 100.0 & 99.6 & 98.8 & 100.0 & 82.6 \\
			\hline
			SpeedReducer & 1 & 100.0 & 80.0 & 100.0 & 90.0 & 80.0 \\
			& 5 & 100.0 & 80.0 & 98.0 & 100.0 & 70.0 \\
			& 10 & 94.0 & 73.0 & 92.0 & 94.0 & 70.0 \\
			& 20 & 96.0 & 78.0 & 93.0 & 96.0 & 72.5 \\
			& 50 & 96.0 & 80.2 & 94.4 & 96.0 & 74.8 \\
			\hline
			ThreeTruss & 1 & 80.0 & 50.0 & 60.0 & 0.0 & 50.0 \\
			& 5 & 68.0 & 50.0 & 60.0 & 8.0 & 54.0 \\
			& 10 & 75.0 & 72.0 & 76.0 & 53.0 & 72.0 \\
			& 20 & 74.5 & 73.0 & 75.0 & 69.5 & 73.0 \\
			& 50 & 75.8 & 76.2 & 76.0 & 74.6 & 76.2 \\
			\hline
			Car & 1 & 100.0 & 100.0 & 80.0 & 100.0 & 50.0 \\
			& 5 & 94.0 & 94.0 & 66.0 & 94.0 & 28.0 \\
			& 10 & 97.0 & 94.0 & 84.0 & 97.0 & 59.0 \\
			& 20 & 96.5 & 94.0 & 86.5 & 96.5 & 73.0 \\
			& 50 & 95.6 & 94.0 & 86.6 & 95.6 & 83.0 \\
			\hline
		\end{tabular}
	\end{minipage}
\end{table*}

\begin{table*}[t]
	\centering
	\caption{Single $\theta$ comparison for all problems ($N=1$). PCM denotes the directly predicted solution from the parametric constraint model, while Refinement represents the locally optimized solution initialized by PCM. Red values indicate infeasible solutions, and bold values denote the best objective among feasible solutions.}
	\label{tab:single_theta_N1}
	\renewcommand{\arraystretch}{1.4}
	\setlength{\tabcolsep}{8pt}
	\begin{tabular}{@{}lccccccc@{}}
		\hline
		Problem & PCM & Refinement & CEI & LogCEI & ALBO & SCBO & CMAES \\
		\hline
		CEC2006-g01 & -1.09e+02 & -1.33e+02 & -1.33e+02 & \textbf{-1.40e+02} & -1.37e+02 & -1.30e+02 & -6.33e+01 \\
		CEC2006-g06 & \textbf{-3.22e+04} & \textbf{-3.22e+04} & {-3.22e+04} & {-3.22e+04} & {-3.22e+04} & -3.22e+04 & -3.22e+04 \\
		CEC2006-g07 & -2.17e+01 & -2.17e+01 & 5.66e+01 & -2.32e+01 & 6.18e+01 & \textbf{-2.35e+01} & 4.18e+01 \\
		CEC2006-g10 & 2.10e+03 & \textbf{2.10e+03} & 2.19e+03 & {2.10e+03} & 2.11e+03 & 2.11e+03 & 3.57e+03 \\
		CEC2010-c13 & -5.02e+01 & -5.02e+01 & -5.22e+01 & -6.97e+01 & \textcolor{red}{5.79e+01} & \textbf{-8.64e+01} & \textcolor{red}{3.76e+01} \\
		Ackley-10D & 4.52e+00 & 4.52e+00 & 3.45e+00 & 4.32e+00 & 6.09e+00 & \textbf{1.99e+00} & 4.27e+00 \\
		Ackley-2D & 3.06e-02 & 3.06e-02 & 1.14e+00 & 4.12e-01 & 2.23e+00 & 8.79e-03 & \textbf{2.29e-03} \\
		Ackley-6D & 3.08e+00 & 3.08e+00 & 2.85e+00 & 3.53e+00 & 5.82e+00 & \textbf{1.13e+00} & 1.34e+00 \\
		Car & 1.54e+01 & \textbf{1.52e+01} & {1.52e+01} & {1.52e+01} & {1.52e+01} & 1.52e+01 & 1.59e+01 \\
		GKXWC1 & -2.00e+00 & -2.00e+00 & -2.00e+00 & -2.00e+00 & \textbf{-2.00e+00} & -2.00e+00 & {-2.00e+00} \\
		GKXWC2 & -1.00e+00 & -1.00e+00 & -1.00e+00 & \textbf{-1.00e+00} & -1.00e+00 & -1.00e+00 & {-1.00e+00} \\
		KeaneBump & -1.86e-01 & -1.86e-01 & -2.13e-01 & {-2.27e-01} & -2.09e-01 & -2.26e-01 & \textbf{-2.30e-01} \\
		PressureVessel & 4.97e+02 & 4.97e+02 & 4.97e+02 & \textbf{4.87e+02} & 5.57e+02 & 5.01e+02 & 8.78e+03 \\
		RCBeam & 1.04e+02 & \textbf{8.99e+01} & 9.45e+01 & {8.99e+01} & 9.01e+01 & 9.00e+01 & 8.99e+01 \\
		SpeedReducer & \textbf{2.35e+03} & \textbf{2.35e+03} & 2.47e+03 & {2.35e+03} & 2.43e+03 & 2.36e+03 & 2.39e+03 \\
		ThreeTruss & \textcolor{red}{2.37e+02} & \textcolor{red}{2.37e+02} & 2.52e+02 & \textbf{2.25e+02} & 2.25e+02 & 2.25e+02 & 2.25e+02 \\
		\hline
		Win Count 
		& (2) & 5 & 0 & 4 & 1 & 4 & 2 \\
		\hline
	\end{tabular}
	
\end{table*}

\begin{table*}[t]
	\centering
	\caption{Single $\theta$ comparison for all problems ($N=5$).}
	\label{tab:single_theta_N5}
	\renewcommand{\arraystretch}{1.4}
	\setlength{\tabcolsep}{8pt}
	\begin{tabular}{@{}lccccccc@{}}
		\hline
		Problem & PCM & Refinement & CEI & LogCEI & ALBO & SCBO & CMAES \\
		\hline
		CEC2006-g01 & \textcolor{red}{-5.74e+01} & -5.37e+01 & \textcolor{red}{-5.84e+01} & \textbf{-5.94e+01} & \textcolor{red}{-4.66e+01} & -5.06e+01 & -3.73e+01 \\
		CEC2006-g06 & \textbf{-3.22e+04} & \textbf{-3.22e+04} & {-3.22e+04} & {-3.22e+04} & {-3.22e+04} & -3.22e+04 & -3.21e+04 \\
		CEC2006-g07 & \textbf{-2.27e+01} & \textbf{-2.27e+01} & 7.95e+01 & -2.04e+01 & 2.42e+02 & -1.09e+01 & 4.06e+02 \\
		CEC2006-g10 & 2.10e+03 & \textbf{2.10e+03} & 2.32e+03 & {2.10e+03} & 6.61e+03 & 2.46e+03 & 8.54e+03 \\
		CEC2010-c13 & \textbf{-4.51e+01} & \textbf{-4.51e+01} & \textcolor{red}{-2.79e+01} & \textcolor{red}{-7.58e+01} & -1.30e+01 & -3.71e+01 & \textcolor{red}{3.77e+01} \\
		Ackley-10D & 4.47e+00 & 4.47e+00 & 6.35e+00 & 4.55e+00 & 7.55e+00 & \textbf{4.35e+00} & 8.70e+00 \\
		Ackley-2D & \textbf{3.04e-02} & \textbf{3.04e-02} & 1.14e+00 & 1.11e-01 & 1.74e+00 & 3.89e-01 & 2.08e+00 \\
		Ackley-6D & \textbf{3.09e+00} & \textbf{3.09e+00} & 3.66e+00 & 3.70e+00 & 5.73e+00 & 3.38e+00 & 4.60e+00 \\
		Car & \textcolor{red}{1.91e+01} & 1.92e+01 & 2.20e+01 & \textbf{1.92e+01} & 1.92e+01 & 2.06e+01 & 2.29e+01 \\
		GKXWC1 & -2.00e+00 & -2.00e+00 & -2.00e+00 & -2.00e+00 & -2.00e+00 & -2.00e+00 & \textbf{-2.00e+00} \\
		GKXWC2 & -9.96e-01 & -1.00e+00 & \textbf{-1.00e+00} & -1.00e+00 & -1.00e+00 & -1.00e+00 & -9.97e-01 \\
		KeaneBump & -1.86e-01 & -1.86e-01 & -1.86e-01 & -1.86e-01 & \textbf{-1.87e-01} & -1.29e-01 & -1.83e-01 \\
		PressureVessel & 5.00e+02 & 5.00e+02 & 5.02e+02 & \textbf{4.92e+02} & 9.37e+02 & 6.70e+02 & 8.78e+03 \\
		RCBeam & 1.05e+02 & 9.14e+01 & 9.62e+01 & \textbf{9.11e+01} & 1.20e+02 & 9.11e+01 & 1.02e+02 \\
		SpeedReducer & \textbf{2.35e+03} & \textbf{2.35e+03} & 2.37e+03 & {2.35e+03} & 2.40e+03 & 2.41e+03 & 2.85e+03 \\
		ThreeTruss & 2.26e+02 & 2.26e+02 & 2.25e+02 & 2.20e+02 & 2.25e+02 & \textbf{2.12e+02} & 2.16e+02 \\
		\hline
		Win Count
		& (6) & 7 & 1 & 4 & 1 & 2 & 1 \\
		\hline
	\end{tabular}
\end{table*}

\begin{table*}[t]
	\centering
	\caption{Single $\theta$ comparison for all problems ($N=10$).}
	\label{tab:single_theta_N10}
	\renewcommand{\arraystretch}{1.5}
	\setlength{\tabcolsep}{8pt}
	\begin{tabular}{@{}lccccccc@{}}
		\hline
		Problem & PCM & Refinement & CEI & LogCEI & ALBO & SCBO & CMAES \\
		\hline
		CEC2006-g01 & \textcolor{red}{-1.80e+02} & -1.31e+02 & -1.31e+02 & \textbf{-1.37e+02} & -1.33e+02 & -1.27e+02 & -1.12e+02 \\
		CEC2006-g06 & \textbf{-3.22e+04} & \textbf{-3.22e+04} & {-3.22e+04} & {-3.22e+04} & {-3.22e+04} & -3.21e+04 & -3.21e+04 \\
		CEC2006-g07 & \textcolor{red}{-2.17e+01} & \textbf{-1.56e+01} & 8.68e+01 & 8.19e+00 & 2.42e+02 & 6.51e+01 & 6.10e+02 \\
		CEC2006-g10 & \textbf{2.10e+03} & \textbf{2.10e+03} & 2.21e+03 & {2.10e+03} & 6.61e+03 & 3.09e+03 & 1.24e+04 \\
		CEC2010-c13 & \textbf{-2.49e+01} & \textbf{-2.49e+01} & \textcolor{red}{2.50e+01} & \textcolor{red}{-8.86e+00} & \textcolor{red}{-2.65e+01} & \textcolor{red}{3.54e+01} & \textcolor{red}{6.43e+01} \\
		Ackley-10D & \textbf{4.46e+00} & \textbf{4.46e+00} & 6.42e+00 & 7.51e+00 & 8.67e+00 & 6.49e+00 & 1.04e+01 \\
		Ackley-2D & \textbf{3.26e-02} & \textbf{3.26e-02} & 1.14e+00 & 4.12e-01 & 1.74e+00 & 3.89e-01 & 2.82e+00 \\
		Ackley-6D & \textbf{3.07e+00} & \textbf{3.07e+00} & 3.67e+00 & 3.75e+00 & 5.73e+00 & 6.07e+00 & 5.34e+00 \\
		Car & 1.52e+01 & \textbf{1.52e+01} &{1.52e+01} & {1.52e+01} & {1.52e+01} & 1.64e+01 & 2.15e+01 \\
		GKXWC1 & \textbf{-2.00e+00} & \textbf{-2.00e+00} & -2.00e+00 & -2.00e+00 & -2.00e+00 & -2.00e+00 & -1.99e+00 \\
		GKXWC2 & -1.00e+00 & -1.00e+00 & \textbf{-1.00e+00} & -1.00e+00 & -1.00e+00 & -9.94e-01 & -9.69e-01 \\
		KeaneBump & \textbf{-1.86e-01} & \textbf{-1.86e-01} & -1.44e-01 & -1.67e-01 & -1.43e-01 & -1.29e-01 & -1.69e-01 \\
		PressureVessel & \textbf{5.02e+02} & \textbf{5.02e+02} & 5.14e+02 & 5.04e+02 & 1.27e+03 & 1.89e+03 & 8.78e+03 \\
		RCBeam & 2.38e+02 & 2.25e+02 & 2.36e+02 & \textbf{2.24e+02} & 2.30e+02 & 2.29e+02 & 2.47e+02 \\
		SpeedReducer & \textcolor{red}{2.35e+03} & \textcolor{red}{2.35e+03} & 3.12e+03 & \textbf{2.65e+03} & 3.21e+03 & 2.79e+03 & 3.89e+03 \\
		ThreeTruss & \textcolor{red}{4.85e+01} & \textbf{4.59e+01} & 5.07e+01 & 6.53e+01 & 6.94e+01 & 4.79e+01 & 6.66e+01 \\
		\hline
		Win Count
		& (9) & 12 & 1 & 3 & 0 & 0 & 0 \\
		\hline
	\end{tabular}
\end{table*}

\begin{table*}[t]
	\centering
	\caption{Single $\theta$ comparison for all problems ($N=20$).}
	\label{tab:single_theta_N20}
	\renewcommand{\arraystretch}{1.5}
	\setlength{\tabcolsep}{8pt}
	\begin{tabular}{@{}lccccccc@{}}
		\hline
		Problem & PCM & Refinement & CEI & LogCEI & ALBO & SCBO & CMAES \\
		\hline
		CEC2006-g01 & \textcolor{red}{-4.76e+01} & -8.01e+01 & \textcolor{red}{-7.45e+01} & \textbf{-9.94e+01} & \textcolor{red}{-4.49e+01} & -8.12e+01 & -3.63e+01 \\
		CEC2006-g06 & \textbf{-3.22e+04} & \textbf{-3.22e+04} & {-3.22e+04} &{-3.22e+04} & {-3.22e+04} & -3.21e+04 & -3.21e+04 \\
		CEC2006-g07 & \textcolor{red}{-2.27e+01} & \textbf{-1.59e+01} & 7.28e+01 & 6.26e+00 & 1.51e+02 & 1.13e+02 & \textcolor{red}{1.96e+03} \\
		CEC2006-g10 & \textbf{2.10e+03} & \textbf{2.10e+03} & 2.23e+03 & {2.10e+03} & 6.61e+03 & 3.29e+03 & 1.31e+04 \\
		CEC2010-c13 & \textbf{-5.73e+01} & \textbf{-5.73e+01} & \textcolor{red}{6.38e+01} & \textcolor{red}{-7.58e+01} & \textcolor{red}{-2.16e+01} & \textcolor{red}{2.56e+01} & \textcolor{red}{6.43e+01} \\
		Ackley-10D & \textbf{4.47e+00} & \textbf{4.47e+00} & 7.07e+00 & 7.86e+00 & 1.02e+01 & 7.85e+00 & 1.14e+01 \\
		Ackley-2D & \textbf{3.14e-02} & \textbf{3.14e-02} & 1.14e+00 & 3.96e-01 & 3.61e+00 & 3.89e-01 & 2.90e+00 \\
		Ackley-6D & \textbf{3.10e+00} & \textbf{3.10e+00} & 3.81e+00 & 4.82e+00 & 7.22e+00 & 6.60e+00 & 6.57e+00 \\
		Car & \textcolor{red}{1.66e+01} & \textbf{1.67e+01} & 1.68e+01 & 1.67e+01 & 2.18e+01 & 1.89e+01 & 2.23e+01 \\
		GKXWC1 & \textbf{-2.00e+00} & \textbf{-2.00e+00} & -2.00e+00 & -2.00e+00 & -2.00e+00 & -1.99e+00 & -1.99e+00 \\
		GKXWC2 & -1.00e+00 & -1.00e+00 & -8.45e-01 & \textbf{-1.00e+00} & -1.00e+00 & -9.89e-01 & -8.31e-01 \\
		KeaneBump & \textbf{-1.86e-01} & \textbf{-1.86e-01} & -1.60e-01 & -1.55e-01 & -1.39e-01 & -1.29e-01 & -1.64e-01 \\
		PressureVessel & \textbf{4.97e+02} & \textbf{4.97e+02} & 5.08e+02 & 5.21e+02 & 1.90e+03 & 3.21e+03 & 8.78e+03 \\
		RCBeam & 8.99e+01 & \textbf{8.99e+01} & {8.99e+01} & {8.99e+01} & {8.99e+01} & 9.43e+01 & 1.08e+02 \\
		SpeedReducer & \textbf{2.35e+03} & \textbf{2.35e+03} & 2.37e+03 & {2.35e+03} & 2.40e+03 & 2.46e+03 & 3.05e+03 \\
		ThreeTruss & 1.42e+02 & 1.42e+02 & 1.49e+02 & 1.43e+02 & 1.65e+02 & 1.47e+02 & \textbf{1.41e+02} \\
		\hline
		Win Count
		& (10) & 13 & 0 & 2 & 0 & 0 & 1 \\
		\hline
	\end{tabular}
\end{table*}

\begin{table*}[t]
	\centering
	\caption{Single $\theta$ comparison for all problems ($N=50$).}
	\label{tab:single_theta_N50}
	\renewcommand{\arraystretch}{1.5}
	\setlength{\tabcolsep}{8pt}
	\begin{tabular}{@{}lccccccc@{}}
		\hline
		Problem & PCM & Refinement & CEI & LogCEI & ALBO & SCBO & CMAES \\
		\hline
		CEC2006-g01 & -6.92e+01 & -9.12e+01 & \textcolor{red}{-1.04e+02} & \textbf{-9.28e+01} & \textcolor{red}{-4.49e+01} & -8.09e+01 & -3.70e+01 \\
		CEC2006-g06 & \textbf{-3.22e+04} & \textbf{-3.22e+04} & {-3.22e+04} & {-3.22e+04} & -3.22e+04 & -3.20e+04 & -3.15e+04 \\
		CEC2006-g07 & \textbf{-2.14e+01} & \textbf{-2.14e+01} & 1.13e+02 & 2.12e+02 & \textcolor{red}{2.81e+03} & 1.17e+02 & \textcolor{red}{1.96e+03} \\
		CEC2006-g10 & \textbf{2.10e+03} & \textbf{2.10e+03} & 2.23e+03 & \textbf{2.10e+03} & 6.61e+03 & 4.11e+03 & 1.31e+04 \\
		CEC2010-c13 & \textbf{-2.36e+01} & \textbf{-2.36e+01} & \textcolor{red}{2.50e+01} & \textcolor{red}{-7.58e+01} & \textcolor{red}{7.03e+01} & \textcolor{red}{2.56e+01} & \textcolor{red}{6.43e+01} \\
		Ackley-10D & \textbf{4.47e+00} & \textbf{4.47e+00} & 7.61e+00 & 9.01e+00 & 1.02e+01 & 1.02e+01 & 1.14e+01 \\
		Ackley-2D & \textbf{3.22e-02} & \textbf{3.22e-02} & 1.14e+00 & 4.12e-01 & 3.71e+00 & 3.89e-01 & 2.90e+00 \\
		Ackley-6D & \textbf{3.09e+00} & \textbf{3.09e+00} & 6.34e+00 & 5.40e+00 & 8.01e+00 & 8.17e+00 & 6.57e+00 \\
		Car & 1.98e+01 & \textbf{1.96e+01} & 2.20e+01 & 1.96e+01 & 1.98e+01 & 2.21e+01 & 2.45e+01 \\
		GKXWC1 & \textbf{-2.00e+00} & \textbf{-2.00e+00} & -1.99e+00 & -2.00e+00 & -1.97e+00 & -1.96e+00 & -1.75e+00 \\
		GKXWC2 & -8.00e-01 & \textbf{-8.23e-01} & -6.10e-01 & -8.07e-01 & -5.60e-01 & -8.16e-01 & -5.93e-01 \\
		KeaneBump & \textbf{-1.86e-01} & \textbf{-1.86e-01} & -1.44e-01 & -1.55e-01 & -1.39e-01 & -1.29e-01 & -1.64e-01 \\
		PressureVessel & \textcolor{red}{5.00e+02} & 3.45e+03 & \textbf{2.86e+03} & 3.92e+03 & 2.94e+04 & 1.77e+04 & 2.03e+04 \\
		RCBeam & 9.60e+01 & \textbf{9.51e+01} & 1.48e+02 & 9.53e+01 & 1.20e+02 & 1.08e+02 & 1.20e+02 \\
		SpeedReducer & \textbf{2.35e+03} & \textbf{2.35e+03} & 2.46e+03 & \textbf{2.35e+03} & 2.40e+03 & 2.54e+03 & 3.05e+03 \\
		ThreeTruss & \textbf{9.87e+01} & \textbf{9.87e+01} & 1.14e+02 & 1.26e+02 & 1.38e+02 & 1.15e+02 & 1.02e+02 \\
		\hline
			Win Count
		& (11) & 14 & 0 & 2 & 0 & 0 & 0 \\
		\hline
	\end{tabular}
\end{table*}

\end{document}